\documentclass{article}

     \PassOptionsToPackage{numbers, compress}{natbib}

\usepackage[final]{neurips_2022}

\usepackage[utf8]{inputenc} %
\usepackage[T1]{fontenc}    %
\usepackage{hyperref}       %
\usepackage{url}            %
\usepackage{booktabs}       %
\usepackage{amsfonts}       %
\usepackage{nicefrac}       %
\usepackage{microtype}      %
\usepackage{xcolor}         %

\usepackage[pdftex]{graphicx}
\usepackage{color}
\usepackage{subcaption}
\usepackage{amssymb,amsmath,amsthm}
\usepackage{mathabx} %
\usepackage{bm}
\newcommand{\RR}{\mathbb{R}}
\newcommand{\CC}{\mathbb{C}}
\newcommand{\KK}{\mathbb{K}}
\newcommand{\NN}{\mathbb{N}}
\newcommand{\ZZ}{\mathbb{Z}}
\newcommand{\TT}{\mathbb{T}}

\newcommand{\dd}{\mathrm{d}}
\newcommand{\eps}{\varepsilon}

\newcommand{\xx}{{\bm{x}}}
\renewcommand{\aa}{{\bm{a}}}
\newcommand{\bb}{{\bm{b}}}
\newcommand{\cc}{{\bm{c}}}
\newcommand{\uu}{{\bm{u}}}
\newcommand{\vv}{{\bm{v}}}

\newcommand{\Sch}{\mathcal{S}}
\newcommand{\calX}{\mathcal{X}}
\newcommand{\subX}{\mathcal{X}_m}
\newcommand{\perpX}{\subX^\perp}

\DeclareMathOperator{\vol}{vol}
\DeclareMathOperator{\lip}{Lip}
\DeclareMathOperator{\id}{Id}

\newcommand{\iprod}[1]{\langle#1\rangle}
\newcommand{\iiprod}[1]{(\!(#1)\!)}

\newcommand{\sfe}{\mathsf{e}}
\newcommand{\hypothesis}{\mathtt{NN}}
\newcommand{\stareq}{\overset{\star}{=}}
\newcommand{\cube}{Q}
\newcommand{\pbP}{P^*\!}

\newcommand{\gpE}{\mathrm{E}}
\newcommand{\gpSE}{\mathrm{SE}}
\newcommand{\gpO}{\mathrm{O}}
\newcommand{\gpSO}{\mathrm{SO}}
\newcommand{\gpU}{\mathrm{U}}

\newcommand{\gpSym}{\mathrm{S}}

\newcommand{\refeq}[1]{\eqref{eq:#1}}

\newcommand{\refapp}[1]{Appendix~\ref{sec:#1}}
\newcommand{\refsec}[1]{\S~\ref{sec:#1}}

\newcommand{\refthm}[1]{Theorem~\ref{thm:#1}}

\newcommand{\refex}[1]{Example~\ref{ex:#1}}

\theoremstyle{plain}
\newtheorem{thm}{Theorem}%

\theoremstyle{definition}
\newtheorem{dfn}{Definition}

\newtheorem{ex}{Example}
\newtheorem{ex*}{Example}

\theoremstyle{remark}

\allowdisplaybreaks

\title{Universality of Group Convolutional Neural Networks Based on Ridgelet Analysis on Groups}

\author{%
  Sho Sonoda\\
  RIKEN AIP\\
  \texttt{\small sho.sonoda@riken.jp}%
  \And
  Isao Ishikawa\\
  Ehime University\\
  \texttt{\small ishikawa.isao.zx@ehime-u.ac.jp}%
  \And
  Masahiro Ikeda\\
  RIKEN AIP\\
  \texttt{\small masahiro.ikeda@riken.jp}
}

\begin{document}

\maketitle

\begin{abstract}
We show the universality of depth-2 group convolutional neural networks (GCNNs) in a unified and constructive manner based on the ridgelet theory. Despite widespread use in applications, the approximation property of (G)CNNs has not been well investigated. The universality of (G)CNNs has been shown since the late 2010s. Yet, our understanding on how (G)CNNs represent functions is incomplete because the past universality theorems have been shown in a case-by-case manner by manually/carefully assigning the network parameters depending on the variety of convolution layers, and in an indirect manner by converting/modifying the (G)CNNs into other universal approximators such as invariant polynomials and fully-connected networks. In this study, we formulate a versatile depth-2 continuous GCNN $S[\gamma]$ as a nonlinear mapping between group representations, and  directly obtain an analysis operator, called the \emph{ridgelet trasform}, that maps a given function $f$ to the network parameter $\gamma$ so that $S[\gamma]=f$. The proposed GCNN covers typical GCNNs such as the cyclic convolution on multi-channel images, networks on permutation-invariant inputs (Deep Sets), and $\mathrm{E}(n)$-equivariant networks. The closed-form expression of the ridgelet transform can describe how the network parameters are organized to represent a function. While it has been known only for fully-connected networks, this study is the first to obtain the \emph{ridgelet transform for GCNNs}. By discretizing the closed-form expression, we can systematically generate a constructive proof of the \emph{$cc$-universality of finite GCNNs}. In other words, our universality proofs are more unified and constructive than previous proofs.
\end{abstract}

\section{Introduction}

In the research field of geometric deep learning \citep{Bronstein2021}, \emph{group convolutional neural  networks (GCNNs)} have been developed to capture the inductive bias behind a variety of datasets
such as sets and point clouds \citep{Qi2017,Zaheer2017}, graphs \citep{Kondor2018,Maron2019graph}, manifolds, groups, and homogeneous spaces \citep{Cohen2018,Cohen2019,Kondor2018a,Kondor2018}.
Despite the rapid growth of diversity, the \emph{approximation property} of CNNs is less investigated than that of fully-connected neural networks (FNNs).
To this date, several authors have shown the universality of (G)CNNs. That is, they can approximate some class of continuous maps with any precision
\citep{Zhou2018,Zhou2020a,Yarotsky2021a,Maron2019universality,Petersen2020,Kumagai2022,okumoto2022learnability}.
These studies are still limited because the proofs are shown
(1) in a case-by-case manner by manually assigning the parameters for a network to approximate a given function $f$, which means that once the network architecture is modified, then we need to reassign the parameters from scratch,
and (2) in an indirect manner by converting/modifying the (G)CNNs into other universal approximators such as invariant polynomials and FNNs, which means that we know only indirectly about (G)CNNs.

The approximation property of \emph{FNNs} has been investigated in the 1990s, with gradually increasing the resolution of proofs from abstract to concrete, starting from purely \emph{existential} proofs based on the Hahn-Banach theorem \citep{Cybenko1989} and the Stone-Weierstrass theorem \citep{Hornik1989}, \emph{indirect} proofs based on the Fourier transform \citep{Irie1988,Funahashi1989}, the Radon transform  \citep{Carroll.Dickinson,Ito.Radon}, $B$-splines \citep{Mhaskar1992,Mhaskar1996}, to more \emph{constructive} proofs based on the integral representation \citep{Barron1993}, ridge functions \citep{Leshno1993}, and the \emph{ridgelet transform} \citep{Murata1996,Candes.PhD,Rubin.calderon}. For deep-ReLU-FNNs, further approximation properties have been investigated \citep{Telgarsky2016,Yarotsky2017,Petersen2018} in the 2010s.
In this context, (G)CNN studies are at the stage of case-by-case and indirect proofs. (See \refsec{related} for more details).

In this study, we show the universality of depth-2 GCNNs by devising a general notion of group convolution and developing the \emph{ridgelet transform for GCNNs}---an analysis operator that maps a given function $f$ to the weight parameter $\gamma$ in a single hidden layer of a neural network. %
Consequently, our universality proof is more \emph{unified and constructive} because our GCNN covers a wide range of typical GCNNs, and the ridgelet transform can describe how to assign the network parameters.

In the following, we describe the formulation of GCNNs to overview our main contributions. %
\paragraph{A Typical Convolution Layer for Images.}
Given an $m_1 \times m_2$-dimensional $n_{in}$-channel input image $\xx\in\RR^{ m_1 \times m_2 \times n_{in}}$, a typical \emph{convolution layer}
with $w_1 \times w_2$-dimensional $n_{in} \times n_{out}$-channel filter $\aa \in \RR^{w_1 \times w_2 \times n_{in}\times n_{out}}$ and $n_{out}$-channel bias $\bb\in\RR^{n_{out}}$ followed by an elementwise activation function $\sigma:\RR\to\RR$ and the aggregation with output coefficients $\cc \in \RR^{n_{out}}$ is given by
\begin{align}
S[\aa,\bb,\cc](\xx)(i,j) = \sum_{\ell=1}^{n_{out}} c^\ell \sigma\left( \sum_{k=1}^{n_{in}} \sum_{p=1}^{w_1}\sum_{q=1}^{w_2} a_{pq}^{k\ell} x_{i+p,j+q}^k - b^\ell \right)
\end{align}
for each pixel at $(i,j) \in [m_1-w_1+1] \times [m_2-w_2+1]$.

For technical reasons, we assume that the output channels (indexed by $\ell \in [n_{out}]$) are aggregated soon after the activation function, which may be slightly different from an ordinary formulation of CNNs, but we can understand this as a part of the subsequent layer.

In the standard formulation of GCNNs, a multi-channel image is understood as a vector-valued function on a group $G$ or a homogeneous space $G/H$, such as a product group $G = \ZZ_{m_1} \times \ZZ_{m_2}$ of cyclic groups $\ZZ_{m_i} := \ZZ / m_i \ZZ \ (i=1,2)$. (More geometrically, \citet{Cohen2017} phrased it as `a section of a fiber bundle'). The convolution in the pixel directions $(i,j)$ is reformulated as a group convolution with respect to the product group, and the inner product in the channel direction $k$ is understood as the convolution with respect to the trivial action of $G$ on a `fiber' $\RR^{n_{in}}$.

\paragraph{The Integral Representation $S[\gamma]$ of Group Convolution Layer.}
Let $G$ be an \emph{arbitrary} group, $\sigma:\RR\to\RR$ be an \emph{arbitrary} nonlinear function, $\calX$ be an \emph{arbitrary} Hilbert space of feature vector $x$ and filter $a$, and $\gamma:\calX\times\RR\to\CC$ be an \emph{arbitrary} function, called the \emph{parameter distribution}.
We formulate a group convolution layer in an integral form, called the \emph{integral representation}, as 
\begin{align}
    S[\gamma](x)(g) := \int_{\calX \times \RR} \gamma(a,b) \sigma\big( (a * x)(g) - b \big) \dd a \dd b, \quad x \in \calX, \ g \in G.
\end{align}
This is an infinite-dimensional reparametrization of a depth-2 GCNN; namely, each function $x \mapsto \sigma( ( a * x)(g) -b)$ represents a single convolutional neuron, or a feature map of input $x$ parametrized by $(a,b)$, the integration over $(a,b)$ means that all the neurons are assigned, and a single function $\gamma$---the \emph{parameter distribution}---parameterizes the assignment of each parameters $(a,b)$. Hence, $S[\gamma]$ can be understood as a \emph{continuous neural network}. We note that, however, if we put $\gamma$ as a finite sum of Dirac's measures such as $\gamma_n := \sum_{\ell=1}^n c^\ell \delta_{(a^\ell,b^\ell)}$, then the integral representation can also represent a finite model
\begin{align}
S[\gamma_n](x)(g) = \sum_{\ell=1}^n c^\ell \sigma( (a^\ell * x)(g)-b^\ell), \quad 
x \in \calX, \ g \in G. \label{eq:finite.model}
\end{align}
In summary, $S[\gamma]$ is a mathematical model of shallow neural networks with \emph{any} width ranging from finite to continuous. In particular, the sparsity/low-rankness of parameters are reflected as the localization/concentration of parameter distribution $\gamma$.

An advantage to use the integral representation
is the \emph{linearization trick}.
Whereas a finite network $S[\gamma_n]$ is \emph{nonlinear} in the original parameters $(a,b)$, the integral representation $S[\gamma]$ is \emph{linear} in the parameter distribution $\gamma$.
It is first emerged in the 1990s to investigate the expressive power of infinitely-wide shallow FNNs \citep{Irie1988,Funahashi1989,Carroll.Dickinson,Ito.Radon,Barron1993, Murata1996, Candes.PhD, Rubin.calderon}; and it is as well common in today's deep learning theory, for example, to investigate the learning dynamics of SGD such as neural tangent kernel (NTK) \citep{Jacot2018,Lee2019,Arora2019b}, lazy learning \citep{Chizat2019}, lottery tickets \citep{Frankle2019}, mean field theory \citep{Nitanda2017,Mei2018,Rotskoff2018,Chizat2018,Sirignano2020}, and Langevin dynamics \citep{Suzuki2020}.

\paragraph{The Ridgelet Transform $R[f;\rho](a,b)$}
is a right inverse (or pseudo-inverse) operator of the integral representation operator $S$. As an outcome of this study, we have obtained its closed-form expression:
\begin{align}
R[f;\rho](a,b) := \int_{\calX}f(x)(e) \overline{\rho( \iprod{a,x}_{\calX} - b )}\dd a \dd b, \quad (a,b) \in \calX \times \RR,    
\end{align}%
where $f : \calX \to \CC^G$ is a target vector-valued nonlinear function to be approximated, called a \emph{feature map}, $e \in G$ is the identity element, and $\rho:\RR\to\CC$ is an auxiliary function, called the \emph{ridgelet function}. 
Provided that $f$ is \emph{group equivariant}, then under mild regularity assumptions, it satisfies the \emph{reconstruction formula}
\begin{align}
    S[R[f;\rho]] = \iiprod{\sigma,\rho} f,
\end{align}
where $\iiprod{\cdot,\cdot}$ denote a scalar product of $\sigma$ and $\rho$. 
Therefore, as long as the product $\iiprod{\sigma,\rho}$ is neither $0$ nor $\infty$, we can normalize $\rho$ to satisfy $\iiprod{\sigma,\rho}=1$ so that $S[R[f;\rho]]=f$. 

In other words, $R$ and $S$ are \emph{analysis and synthesis operators}, and thus play the same roles as the Fourier ($F$) and inverse Fourier ($F^{-1}$) transforms, respectively. Particularly, the reconstruction formula $S[R[f;\rho]]=\iiprod{\sigma,\rho}f$ corresponds to the Fourier inversion formula $F^{-1}[F[f]]=f$. 

An advantage of the ridgelet transform is the \emph{closed-form expression}.
Despite the common belief that neural network parameters are a blackbox, the closed-form expression can clearly describe how the network parameters are organized.
Previous studies on the CNN universality have also provided several construction algorithms of parameters, but these are only \emph{particular} solutions for a CNN to represent a target function $f$, and not necessary related to, for example, deep learning solutions.
For \emph{FNNs}, on the other hand, \citet{Sonoda2021ghost} have shown that any parameter distribution $\gamma$ satisfying $S[\gamma]=f$ can always be represented as (not always single but) a linear combination of ridgelet transforms, and they \citep{Sonoda2021aistats} have shown that finite networks trained by regularized empirical risk minimization (RERM) converges to a certain unique ridgelet transform.
(We note that NTK and the Gibbs distribution can also describe the parameter distribution, but NTK is limited to the kernel regime, and the Gibbs distribution is given only implicitly.)
As an application, \citet{Savarese2019} and their followers \citep{Ongie2020,Parhi2021,Unser2019} 
have established the \emph{representer theorems} for ReLU-FNNs by using the ridgelet transform. 
Although the parallel results for CNNs have not yet been published, we anticipate that the ridgelet transform could facilitate our understanding of deep learning solutions.

\paragraph{Challenges and Contributions.}
The closed-form expression of the ridgelet transform has been known only for FNNs, which  was discovered in the 1990s independently by \citet{Murata1996}, \citet{Candes.PhD} and \citet{Rubin.calderon}.
 (We refer to \citep{Donoho2002,Starck2010,Kutyniok2012} for ridgelet analysis in the 2000s, and \citep{Sonoda2015acha,Kostadinova2014,sonoda2022symmetric} for more recent results.)
One of the difficulties to obtain the ridgelet transform for CNNs is that there is no unique way to formulate an ``integral representation of CNNs''. 
We note that some authors claim the ``equivalence of CNNs and FNNs'' (see e.g. \citep{Petersen2020}), but it is somewhat misleading because such an equivalence holds only when both CNNs and FNNs are very carefully designed.
While FNNs are defined on the Euclidean space $\RR^m$, GCNNs are defined on a more abstract space $\calX$.
For example, since the convolution on the Euclidean space can be written using T\"oplitz matrices, one could consider a formulation such as $\int_{\RR^{k \times m} \times \RR^k} \gamma(A,b) \sigma(Ax-b) \dd A \dd b$ where the parameter $A$ is an $k \times m$-matrix. However, this only leads to another ridgelet transform that covers less symmetries $G$.  %
In fact, it is a version of the so-called $k$-plane ridgelet transform developed in the 2000s 
\citep{Donoho2002}.

To circumvent this difficulty, we formulate GCNNs as general as possible by dealing with the feature space $\calX$, group $G$, and representation $T$ in a coordinate-free manner.
Eventually, we have shown the reconstruction formula for a wide range of GCNNs (as displayed in \refsec{examples}), with a relatively \emph{simple} proof. This study is the first to obtain the ridgelet transform for a general class of GCNNs. As an application, we show the $cc$-universality of GCNNs for a general class of group equivalent continuous vector-valued functions in a unified and constructive manner.

\section{Notation and Basic Terminologies}

\paragraph{Notation.}
For any integer $n>0$, $[n]$ denotes the set $\{ 1, \ldots, n \}$. For any sets $G$ and  $\KK$, $\KK^G$ denotes the collection $\{ G \to \KK \}$ of all mappings from $G$ to $\KK$.
For any topological space $X$, 
$C(X)$ and $C_c(X)$ denote the collections of all continuous functions on $X$, and continuous functions on $X$ with compact support, respectively. We note that when $X$ is compact, then $C(X) = C_c(X)$.
For any measure space $X$ and number $p \in [1,\infty]$, $L^p(X)$ denotes the space of $p$-integrable functions on $X$.

\subsection{Fourier Analysis on $\RR^d$} \label{sec:fourier}
We refer to \citep{Grafakos.classic,GelfandShilov,Sonoda2015acha} for more details on Fourier transform and tempered distributions ($\Sch'$).

\paragraph{Schwartz Distributions.}
For any integer $d>0$, $\Sch(\RR^d)$ and $\Sch'(\RR^d)$ denote the classes of Schwartz test functions (or rapidly decreasing functions) and tempered distributions on $\RR^d$, respectively. Namely, $\Sch'$ is the topological dual of $\Sch$. In this study, $\Sch'(\RR)$ and $\Sch(\RR)$ are assigned as classes of activation and ridgelet functions, respectively. We note that $\Sch'(\RR)$ includes truncated power functions $\sigma(b)=b_+^k = \max\{b,0\}^k$ such as step function for $k=0$ and ReLU for $k=1$. %

\paragraph{Fourier Transform.}
The Fourier transform on the Euclidean space $\RR^d$ and its inversion formula has been defined on (at least) three different function classes: $L^1(\RR^d), L^2(\RR^d)$ and $\Sch'(\RR^d)$. When $f \in L^1(\RR^d)$ and $\widehat{f} \in L^1(\RR^d)$, the inversion formula holds ``at every continuous point $\xx$ of $f$'', which is a pointwise equation. When $f \in L^2(\RR^d)$, the inversion formula holds ``in $L^2$'', which is not a pointwise equation because the equation ``$f=g$ in $L^2$'' is defined as ``$f(\xx)=g(\xx)$ a.e.''. Similarly, when $f \in \Sch'(\RR^m)$, the inversion formula holds ``in $\Sch'$''. We use the third definition for computing the Fourier transform of activation functions $\sigma \in \Sch'(\RR)$ such as ReLU and $\tanh$.

\subsection{Group Representation}
Let $G$ be a group, let $\calX$ be a vector space over a field $\KK$, and let $GL(\calX)$ be the general linear group on $\calX$. A \emph{group representation} $T$ of the group $G$ on the vector space $\calX$ is a group homomorphism from $G$ to $GL(\calX)$, that is, a map $T:G\to GL(\calX); g \mapsto T_g$ satisfying $T_{gh} = T_g T_h$ for all $g,h \in G$. 
When $G$ is a topological group, we further assume that the action $G \times \calX \to \calX; (g,x) \mapsto T_g[x]$ be continuous.
Here, $\calX$ is called the \emph{representation space}. %
We refer to \citep{Folland2015} for more details on group representation.

\paragraph{Regular Representation.}
Let $\calX$ be the vector space of all functions on $G$, i.e., $\calX = \KK^G$.
The \emph{(left) regular representation} $L$ is a group representation defined on $\calX$ as
\begin{align}
    L_g[x](h) := x(g^{-1}h), \quad g,h \in G, \ x \in \calX = \KK^G.
\end{align}
In particular, when $G$ is a locally compact Haussdorf (LCH) group, then it has a (left) invariant measure $\mu$, and we can define the collection $L^2(G)$ of all square integrable functions on $G$ with respect to the canonical inner product
    $\iprod{x,y}_{L^2(G)} := \int_{G} x(g)\overline{y(g)} \dd \mu(g)$ for any measurable functions $x,y:G \to \CC$.
It is known that the regular representation on $\calX = L^2(G)$ is a unitary representation.

\paragraph{Dual Representation.} 
For any group representation $T:G\to GL(\calX)$, the \emph{dual representation} $T^*:G\to GL(\calX')$ is a group representation defined on the dual vector space $\calX'$ as the transpose of $T_{g^{-1}}$, that is, $T_g^* = T_{g^{-1}}^\top$. %
When $\calX$ is a Hilbert space with inner product $\iprod{\cdot,\cdot}_{\calX}$, then it satisfies the following relation:
\begin{align}
    \iprod{T_g[x],y}_{\calX} = \iprod{x,T_{g^{-1}}^*[y]}_{\calX}, \quad g \in G, \ x,y \in \calX.
\end{align}

\paragraph{Matrix Element.} For any group representation $T:G\to GL(\calX)$, the \emph{matrix element} (or the \emph{matrix coefficient}) of $T$ is a bilinear functional $f_{a,x}$ on $G$ defined by 
\begin{align}
f_{a,x}(g) := a[ T_g[x] ], \quad g \in G, \ x \in \calX, \ a \in \calX'
\end{align}
where $x$ is a vector in $\calX$ and $a \in \calX'$ is a continuous linear functional on $\calX$. When $\calX$ is a Hilbert space, then (identifying $\calX'$ with $\calX$) it can be written as 
\begin{align}
    f_{a,x}(g) = \iprod{T_g[x],a}_{\calX}, \quad g \in G, \ a,x \in \calX.
\end{align}
In the next section, we use this quantity as the generalized form of the \emph{group convolution}.

\subsection{Universality} 
The notion of \emph{universality} in machine learning can be rephrased as the \emph{density} in mathematics, and thus it has several definitions. (See, e.g., \citep{Sriperumbudur2010,Pinkus.survey}). In this study, we show the so-called \emph{$cc$-univesality}, one of the standard universalities in the machine learning theory. 

\paragraph{$cc$-Universality.}
Let $X$ be a topological space, and let $\hypothesis$ be a collection of functions (e.g., neural networks) on $X$.
The $cc$-univesality of $\hypothesis$ is defined as the density of $\hypothesis$ in $C(X)$ endowed with the topology of \emph{compact convergence}, that is, for any compact subset $K \subset X$, continuous function $f \in C(K)$, and all $\eps > 0$, there exists a function $g \in \hypothesis$ %
such that
\begin{align}
\big\| f - g|_K \big\|_{C(K)} := \sup_{x \in K} | f(x) - g(x) | < \eps,
\end{align}
where $g|_K$ denotes the restriction of $g$ to $K$.

\section{Functions on Abstract Hilbert Space $\calX$}
We introduce an extended group convolution on $\calX$, a uniform norm and the group-equivariance for functions on $\calX$, an induced measure and an induced Fourier transform on $\calX$, and a projection to $\subX$.

\subsection{$(G,T)$-Convolution $*_T : \calX \times \calX \to \CC^G$}%
\begin{dfn}%
    Let $G$ be a group,
    let $\calX$ be a %
    Hilbert space with inner product $\iprod{\cdot,\cdot}_{\calX}$ over a field $\KK$,
    and let $T:G\to GL(\calX)$ be a representation of $G$ on $\calX$.
    For any $a,x \in \calX$ and $g \in G$, we define the \emph{$(G,T)$-convolution} as
    \begin{align}
        (a *_T x)(g) := \iprod{x,T_g^*[a]}_{\calX} = \iprod{T_{g^{-1}}[x],a}_{\calX}.
    \end{align}
\end{dfn}

We remark (1) that this is simply a paraphrase of the \emph{matrix element of a group representation} (see the previous section), and (2) that this is \emph{not necessarily} a binary operation because $\calX \neq \CC^G$ in general. Nevertheless, we call it a \emph{convolution} simply because it covers a wide range of `group convolutions' in today's GCNN literature.

\begin{ex}
An orthodox group convolution is reproduced when $T$ is the regular representation (of a LCH group $G$) on $\calX=L^2(G)$, i.e., $T_g^*[a](h) = a(g^{-1}h)$. In fact, 
\begin{align}
    \iprod{x,T_g^*[a]}_{L^2(G)}
    = \int_{G} x(h)\overline{a(g^{-1}h)}\dd\mu(h)
    = \int_{G} x(h) \widetilde{a}(h^{-1}g)\dd\mu(h)
    = (x *_G \widetilde{a})(g),
\end{align}
where $\widetilde{a}(g) := \overline{a(g^{-1})}$ is an involution.
\end{ex}
\begin{ex}
The cyclic convolution for an $n$-channel image $\xx = (x_{ij}^k)\in \RR^{m_1\times m_2 \times n}$ is understood as the case when $G=\ZZ_{m_1} \times \ZZ_{m_2}$, 
$\calX = \RR^{m_1 \times m_2 \times n}$, and $T_{(p,q)}^*[\aa](i,j,k) = a_{i-p,j-q}^k$, then
\begin{align}
        \iprod{\xx,T_{(p,q)}^*[\aa]}_{\RR^{m_1 \times m_2 \times n}} = \sum_{i,j,k} x_{ij}^k a^k_{i-p,j-q}.
\end{align}
\end{ex}

While the post-activation feature $\sigma( (a*x)(g) - b)$ is a function on $G$,
the input feature $x$ can be an arbitrary abstract vector,
which is more general than typical GCNN formulations where feature $x$ is supposed to be a vector-valued function on $G$ or $G/H$. This is an advantage for a more geometric understanding of CNNs, since the theory becomes free from the specification of $x$.

\subsection{Continuous $(G,T)$-Equivariant Vector-Valued Function $f:\calX \to C(G)$}
\begin{dfn}%
We say a vector-valued function $f:\calX \to \CC^G$ is \emph{$(G,T)$-equivariant} when
\begin{align}
    f( T_g[x] )(h) = L_{g}[f(x)](h) = f(x)(g^{-1}h), \quad x \in \calX, \ g,h \in G.
\end{align}
\end{dfn}
Here, we restrict the definition for a special case of the regular representation $L$. This is simply due to the fact that our GCNN satisfies this case. %

\begin{dfn}
Let $G$ be a topological group. For any vector-valued function $f:\calX \to \CC^G$, put
\begin{align}
    \| f \|_{C(\calX; C(G))} := \| f \|_{C(\calX) \to C(G)} :=  \sup_{x \in \calX} \big|  \sup_{g \in G} | f(x)(g) | \big|.
\end{align}
By $C_{equi}(\calX;C(G))$, we denote the normed vector space of all \emph{continuous $(G,T)$-equivariant $C(G)$-valued} functions on $\calX$ equipped with the uniform norm $\| \cdot \|_{C(\calX;C(G))}$. %
\end{dfn}
We note that the topology of uniform norm $\| \cdot \|_{C(\calX;C(G))}$ is stronger than the topology of compact convergence, which is employed in the $cc$-universality argument. In fact, $C_{equi}(\calX;C(G))$ need not be complete (or Banach) to show the $cc$-universality.

\subsection{Induced Lebesgue Measure $\lambda$ and Induced Fourier Transform $\widehat{\cdot}\,$ on Subspace $\subX$}
Let $\subX$ denote an $m$-dimensional subspace of $\calX$, and let $\{ \sfe_i \}_{i \in [m]}$ be an orthonormal basis of $\subX$.

    \paragraph{Induced Lebesgue Measure on $\subX$.}
    We induce the Lebesgue measure $\lambda$ on $\subX$ by pushing forward the Lebesgue measure $\dd\xx$ on $\RR^m$ via an isometric linear embedding $\phi:\RR^m\to\subX$.
    For example, take a linear embedding $\phi(\xx) := \sum_{i \in [m]} x_i \sfe_i$. Then, it preserves the length, and we can induce the Lebesgue measure $\lambda$ on $\subX$ as the push forward measure $\lambda = \phi_\sharp \dd\xx$ so that the volume of a hypercube 
    $\cube = \{ \sum_{i \in [m]} c_i \sfe_i \mid c_i \in [a_i,b_i] \}$ in $\subX$
    is calculated as 
    $\lambda(\cube) = \prod_{i \in [m]} | b_i - a_i |$, and the integration of a measurable function $f:\subX\to\CC$ over a measurable set $E \subset \subX$ is calculated as
    \begin{align}
    \int_{E} f( x ) \dd \lambda(x) = \int_{\RR^m} 1_E\left( \sum_{i=1}^m x_i \sfe_i \right) f\left( \sum_{i=1}^m x_i \sfe_i \right) \prod_{i=1}^m \dd x_i = \int_{\phi^{-1}(E)} f\circ \phi(\xx)\dd\xx.
    \end{align}
    As far as there is no risk of confusion, we denote $\dd x$ instead of $\dd \lambda(x)$.

    \paragraph{Induced Fourier Transform on $\subX$.}
    Using $\lambda$, we induce the Fourier transform on $\RR^m$ as below: For any function $f:\subX\to\CC$,
    \begin{align}
        \widehat{f}(y) := \int_{\subX} f(x) e^{-i\iprod{x,y}_{\subX}} \dd \lambda(x), \quad 
        f(x) \overset{\star}{=} \frac{1}{(2\pi)^m} \int_{\subX} \widehat{f}(y) e^{i\iprod{x,y}_{\subX}} \dd \lambda(y).
    \end{align}
    Here, the equality $\overset{\star}{=}$ holds in at least three different senses (see the comments in \refsec{fourier}).

    We remark (1) that once the subspace $\subX$ is fixed, the induced Fourier transform is unique up to the orthogonal transformation of the basis $\{\sfe_i\}_{i \in [m]}$, and (2) that the induced Fourier transform ``on $\calX$'' should not be confused with the Fourier transform ``on group $G$''. Especially, this \emph{cannot} map a convolution $x *_T a$, an element in $\CC^G$, to a point product such as ``$\widehat{x} \cdot \widehat{a}$''.

\subsection{Projection $P:\calX\to\subX$ and Extension Operator $\pbP$}
In order to induce the Lebesgue measure $\lambda$, we assume that the dimension of $\subX$ to be finite. As a side effect of this assumption, the image $T_G[\subX] := \{ T_g[x] \mid g \in G, x \in \subX \}$ can extend toward the outside of $\subX$; that is, $\subX$ is not necessarily $G$-invariant ($T_G[\subX] \subset \subX$). To avoid an ``undefined error'' such as to input $x$ outside of $\subX$ for a function $f$ defined only on $\subX$, we introduce projection $P$ and extension $\pbP$ as below. When $\dim \calX < \infty$, we can omit $P$ by putting $\subX=\calX$ (so $P=\id$), because by the definition of the group representation, always $T_G[\calX]=\calX$.

Let $\perpX$ denote the orthogonal complement of $\subX$ in $\calX$.
Let $P:\calX\to\subX$ denote the orthogonal projection onto $\subX$.
For any function $f:\subX\to\CC^G$, put
\begin{align}
\pbP f(z)(g) := f(P(z))(g), \quad z \in \calX, g \in G.
\end{align}
This extends $f$ (on a subspace $\subX$) to the entire space $\calX$ as a constant function on $\perpX$; that is, $\pbP f( x \oplus y ) = f(x \oplus 0)$ for each $x \oplus y \in \subX \oplus \perpX$.

\section{Main Results}
We introduce the $(G,T)$-convolutional neural networks and the corresponding ridgelet transform, and present the reconstruction formula for continuous GCNNs and the $cc$-universality for finite GCNNs.

Throughout this section, we fix a representation $T:G\to GL(\calX)$ of a group $G$ on a (potentially infinite-dimensional) Hilbert space $\calX$ over a field $\KK$ endowed with an inner product $\iprod{\cdot,\cdot}_{\calX}$,
and fix an $m$-dimensional closed subspace $\subX$ of $\calX$ equipped with an induced Lebesgue measure $\lambda$.
Let $k := \dim_\RR \KK$ denote the real dimension of $\KK$, that is, $k=1$ for $\KK=\RR$ and $k=2$ for $\KK=\CC$. Let $e$ denote the identity element of $G$.

\subsection{Integral Representation of $(G,T)$-Convolutional Neural Network}
\begin{dfn}
For any functions $\gamma:\subX\times\KK \to \CC$ and $\sigma:\KK\to\CC$, we define the \emph{integral representation of $(G,T)$-convolutional neural network} as a vector-valued function $\calX \to \CC^G$,
\begin{align}
    S[\gamma](x)(g) &:= \int_{\subX\times\KK} \gamma(a,b)\sigma( (a *_T x)(g) - b ) \dd \lambda(a) \dd b, \quad x \in \calX, \ g\in G.
\end{align}
Here, we call $\gamma$ a parameter distribution, and $\sigma$ an activation function.
If there is no risk of confusion, we abbreviate $\dd \lambda(a)$ as $\dd a$.
\end{dfn}

It is easy to see that a $(G,T)$-CNN is $(G,T)$-equivariant. In fact, for every $g,h \in G$,
\begin{align}
    S[\gamma](T_g[x])(h)
    &= \int_{\subX\times\KK} \gamma(a,b)\sigma( \iprod{T_{(g^{-1}h)^{-1}}[x],a}_{\calX} - b ) \dd a \dd b = S[\gamma](x)(g^{-1}h).
\end{align}
In addition, at the identity element $g=e$, it is reduced to a FNN: 
\begin{align}
    S[\gamma](x)(e) = \int_{\subX \times \KK} \gamma(a,b) \sigma( \iprod{x,a}_{\calX} - b ) \dd a \dd b, \quad x \in \calX
\end{align}
and it satisfies a \emph{projection property}:
\begin{align}
    S[\gamma]( P[x] )(e) = S[\gamma](x)(e), \quad x \in \calX.
\end{align}

\subsection{Ridgelet Transform and Scalar Product of Activation Function}
\begin{dfn} For any functions $f:\subX\to\CC^G$ and $\rho:\KK\to\CC$, we define the \emph{ridgelet transform} as 
\begin{align}
    R[f;\rho](a,b) &:= \int_{\subX} f(x)(e) \overline{\rho( \iprod{x,a}_{\calX} - b )} \dd x, \quad (a,b) \in \subX\times\KK.
\end{align}
Here $e$ denotes the identity element of $G$.
\end{dfn}

\begin{dfn} For any tempered distribution $\sigma \in \Sch'(\KK)$ and function $\rho \in \Sch(\KK)$, put a \emph{scalar product} as
\begin{align}
    \iiprod{\sigma,\rho} := (2\pi)^{m-k} \int_{\KK} \sigma^\sharp(\omega) \overline{\rho^\sharp(\omega)} |\omega|^{-m} \dd \omega.
\end{align}
Here, $\cdot^\sharp$ denotes the Fourier transform on $\KK$, which is identified with the Fourier transform on $\RR^k$ with $k = \dim_\RR \KK$. We note that $\sigma^\sharp$ is defined in the sense of tempered distributions.
\end{dfn}

The derivations of the ridgelet transform and the scalar product are clarified in the proof of the reconstruction formula. 
Some readers may notice that the ridgelet transform for GCNN is formally the same as the one for FNNs, and may wonder why inner product $\iprod{x,a}_{\calX}$ instead of group convolution $(a *_T x)(g)$. 
Indeed, this is a consequence of two facts (1) that a group convolution at the identity $e$ is reduced to an inner product: $(a *_T x)(e) = \iprod{x,a}_{\calX}$, and (2) that when $f$ is $(G,T)$-equivariant, then the value $f(x)(g)$ at each $g \in G$ is determined by translating the value $f(x)(e)$ at the identity.

\subsection{Reconstruction Formula, or the Universality of Continuous GCNNs}

We state the first half of our main results.
For $f:\subX\to\CC^G$, we write $f_e(x) := f(x)(e)$ for short.

\begin{thm}[Main Theorem 1/2] \label{thm:gconv.fdim.elem.equiv} %
Given a function $f:\subX\to\CC^G$, 
assume (A1) that $\pbP f:\calX\to\CC^G$ is $(G,T)$-equivariant, i.e.,
\begin{align}
&    \pbP f( T_g[z] )(h) = f(z)(g^{-1}h), \quad \mbox{for every } z \in \calX \mbox{ and } g,h \in G;
\end{align}
and (A2) that $f$ satisfies at least one of the following conditions:
 (A2a) both $f_e$ and $\widehat{f_e}$ are absolute-integrable, i.e., $f_e, \widehat{f_e} \in L^1(\subX)$,
 (A2b) $f_e$ is square-integrable, i.e., $f_e \in L^2(\subX)$, or %
 (A2c) $f_e$ is a tempered distribution, i.e., $f_e \in \Sch'(\subX)$. %
Then, the following reconstruction formula holds:
\begin{align}
    S[R[f;\rho]](x)(g) = \int_{\subX\times\KK} R[f;\rho](a,b) \sigma( (a *_T x)(g) - b ) \dd a \dd b
    \stareq \iiprod{\sigma,\rho}f(x)(g), %
\end{align}
where the equality $\stareq$ holds at every continuous point $x_c$ of $f$ for (A2a), in $L^2$ for (A2b), and  in $\Sch'$ for (A2c), respectively.
\end{thm}
The proof is given in \refapp{proof.reconst}.

\subsection{$cc$-Universality of Finite GCNNs}
Finally, we state the second half of our main results.
Let $\hypothesis$ be the collection of finite GCNNs, that is, 
\begin{align}
\hypothesis := \bigcup_{n \in \NN} \left\{ f_n(x)(g) = \sum_{i=1}^n c_i \sigma( (a_i *_T x)(g) - b_i ) \Bigg| (a_i,b_i,c_i) \in \subX \times \KK \times \CC, i \in [n] \right\}.
\end{align}
Since the reconstruction formula 
$S[\gamma_f]=f$ with $\gamma_f = R[f;\rho]$
holds for an arbitrary function $f$,
we can construct a sequence $\{ f_n \}_{n \in \NN}$ of finite $(G,T)$-CNNs that converges to an arbitrary target function $f$, namely
\begin{align}
    f_n \to f \quad \mbox{ as } \quad n \to \infty,
\end{align}
by discretizing the continuous network $S[\gamma_f]$ and distribution $\gamma_f$ into finite sums
\begin{align}
    f_n(x)(g) := S[\gamma_n](x)(g) = \sum_{i=1}^n c_i \sigma( (a_i *_T x)(g) - b_i ) \quad \mbox{with} \quad \gamma_n := \sum_{i=1}^n c_i \delta_{(a_i,b_i)}
\end{align}
in a `nice' manner so that $\gamma_n \to \gamma_f = R[f;\rho]$ as $n \to \infty$.
This is the primitive idea behind the constructive proof of the following $cc$-universality of finite $(G,T)$-CNNs based on ridgelet analysis.

To state a regularity assumption on the activation function $\sigma$, we introduce the forward difference operator $\Delta_\theta^n$ with difference $\theta>0$, defined as
\begin{align}
    \Delta_\theta^{1}[\sigma](t) := \sigma(t + \theta) - \sigma(t), \quad 
    \Delta_\theta^{n+1}[\sigma](t) := \Delta_\theta^{1} \circ \Delta_\theta^{n}[\sigma](t). 
\end{align}

\begin{thm}[Main Theorem 2/2] \label{thm:gcnn.universality}
For an activation function $\sigma \in \Sch'(\KK)$, assume (A3) that there exist $n \in \NN$ and $\theta>0$ such that $\Delta_\theta^n[\sigma]$ is bounded and Lipschitz continuous.
 Then, $\hypothesis$ is $cc$-universal; that is, for any continuous $(G,T)$-equivariant $C(G)$-valued function $f \in C_{equi}( \subX; C(G) )$, and for any compact sets $K \subset \subX$ and $L \subset G$, there exists a sequence $\{ f_n \}_{n \in \NN} \subset \hypothesis$ of finite GCNNs satisfying
\begin{align}
    \| f - f_n \|_{C(K;C(L))} = \sup_{x \in K} \sup_{g \in L} | f(x)(g) - f_n(x)(g) | \to 0, \quad n \to \infty.
\end{align}
\end{thm}
The proof is given in \refapp{proof.universality}. Here, $f \in C_{equi}(\subX;C(G))$ means that $\pbP f$ is $(G,T)$-equivariant.

\section{Examples} \label{sec:examples}
We display the ridgelet transforms and reconstruction formulas for a few typical GCNNs. Besides, we calculated in Examples \ref{ex:d.fpconv} and \ref{ex:d.cpconv} the ridgelet transforms of a differential filter, which is often reported to be acquired as a feature map in the first layer of deep CNNs for image recognition \citep{Krizhevsky2012,Zeiler2014}.

\subsection{Finite Periodic Convolution Layer}
\begin{ex}[For 1-dimensional periodic signals]\label{ex:fpconv.vector}
The periodic convolution corresponds to the case when $\KK=\RR, G=\ZZ_m \cong [m] = \{ 0, 1, \ldots, m-1\}$, $\calX = L^2(G) \cong \RR^m$ equipped with the inner product $\iprod{x,y} := \frac{1}{m}\sum_{i \in [m]} x_i y_i$, and $T_i[x](j) := x_{j-i}$ thus $(a *_T x)(i) = \frac{1}{m}\sum_{j \in [m]} a_j x_{i+j}$. Therefore, the ridgelet transform and the reconstruction formula are given by
\begin{align*}
    &R[f;\rho](a,b) = \int_{\RR^m} f(\xx)(0) \overline{\rho\left( {\textstyle \frac{1}{m}  \sum_{i \in [m]}} a_i x_i - b \right)}\dd \xx,\\
    &S[R[f;\rho]](x)(i) = \int_{\RR^m\times\RR}R[f;\rho](\aa,b)\sigma\left( {\textstyle\frac{1}{m}\sum_{j\in[m]}} a_j x_{i+j} - b\right)\dd \aa\dd b = \iiprod{\sigma,\rho}f(x)(i).
\end{align*}
\end{ex}

\begin{ex}[For 2-dimensional multi-channel periodic images]\label{ex:fpconv.matrix}
A $2$-dimensional $n$-channel image is identified with a vector-valued function $x : \ZZ_m^2 \to \RR^n$, thus $\calX \cong \RR^{m^2 \times n}$. Let $x_{ij}^k$ denote the $(i,j)$-th component in the $k$-th channel of $x \in \calX$. Let $G = \ZZ_m^2$, and put $T_{(p,q)}[x]_{ij}^k := x_{i-p,j-q}^{k}$. 
Therefore, the ridgelet transform and the reconstruction formula are given by
\begin{align*}
    &R[f;\rho](a,b) = \int_{\RR^{m^2 n}} f(x)_{(0,0)} \overline{\rho\left( {\textstyle\frac{1}{m^2 n} \sum_{k \in [n]} \sum_{i,j \in [m]}} a_{ij}^k x_{ij}^k - b \right)}\dd \xx.\\
    &S[R[f;\rho]](x)_{ij} = \int_{\RR^{m^2 n}\times\RR}R[f;\rho](a,b)\sigma\left( {\textstyle\frac{1}{m^2 n}\sum_{k \in [n]}\sum_{p,q \in [m]}} a_{pq}^k x_{p+i,q+j}^k - b\right)\dd \aa\dd b = \iiprod{\sigma,\rho}f(x)_{ij}.
\end{align*}
\end{ex}

\begin{ex}[Difference operator (with cutoff function)] \label{ex:d.fpconv}
A difference operator on $x:[m]\to\RR$ is given by $x = \sum_{i \in [m]} x_i \delta_i \mapsto f(x) = \sum_{i \in [m]}( x_{i+1}-x_i)\delta_i$, which is $(G,T)$-equivariant: $f( T_k^*[x] )(i) =  T_k^*[x]_{i+1}-T_k^*[x]_i = x_{i+1-k}-x_{i-k} = f( x )(i-k)$. Since $f(x)(0) = x_1 - x_0$,
\begin{align*}
    R\left[ f|_K;\rho \right](a,b) = \int_{\RR^m} (x_1-x_0) 1_K(\xx) \overline{\rho\left( {\textstyle \frac{1}{m} \sum_{i\in[m]} } a_i x_i - b \right)} \dd\xx.
\end{align*}
We note since $\xx \mapsto x_1 - x_0$ is not integrable in $\RR^m$, we restrict $f$ to a compact set $K \subset \RR^m$, and impose the indicator function $1_K$ as an auxiliary cutoff function. 
\end{ex}

\subsection{(Deep Sets) Permutation Equivariant Maps on A Finite Set}
\begin{ex}\label{ex:deepsets}%
Let $\calX = \RR^m, G \le \gpSym_m$ and $T_g[x] = (x_{g^{-1}(1)}, x_{g^{-1}(2)}, \ldots, x_{g^{-1}(m)})$.
Thus $\iprod{a,x}_{\calX} = \frac{1}{m} \sum_{i \in [m]} a_i x_i$, and $(a *_T x)(g) = \frac{1}{m}\sum_{p \in [m]} a_{p} x_{g(p)}$. So,
\begin{align*}
    &R[f;\rho](a,b) = \int_{\RR^{m}} f(\xx)(e) \overline{\rho\left( {\textstyle\frac{1}{m} \sum_{i \in [m]}} a_{i} x_{i} - b \right)}\dd \xx,\\
    &S[R[f;\rho]](x)(g) = \int_{\RR^{m}\times\RR}R[f;\rho](\aa,b)\sigma\left( {\textstyle\frac{1}{m}\sum_{p \in [m]}} a_{p} x_{g(p)} - b\right)\dd \aa\dd b = \iiprod{\sigma,\rho}f(x)(g).
\end{align*}
\end{ex}

\subsection{Continuous Periodic Convolution Layer}
\begin{ex}\label{ex:cpconv}%
Let $G = \TT := \RR/2\pi\ZZ \cong \{ e^{i\theta} \mid \theta \in [-\pi,\pi] \}$ be the 1-dimensional torus group, which is one of the most basic continuous group.
As a consequence of the Fourier series expansion, $L^2(\TT)$ is spanned by $\{ e^{i n\theta} \mid n \in \NN\}$. Hence, we can take $\calX$ to be an $m$-dimensional subspace $\calX := \{ \sum_{|n|< m} x_n e^{in\theta} \mid x_{-n}=x_n \in \RR\}$ equipped with an inner product $\iprod{x,y}_{\calX} := \int_\TT x(\theta)\overline{y(\theta)}\dd\theta$. We note that the constraint $x_n = x_{-n}$ implies $\sum_{|n|< m} x_n e^{in\theta} = \sum_{|n|< m} x_n \cos(n\theta)$ and thus any signal $x \in \calX$ is a bandlimited real-valued continuous signal with each coefficient $x_n$ being the $n$-th frequency spectrum. Put $T_{\alpha}[x](\theta) := x(\theta-\alpha)$, then $(a * x)(\alpha) = \int_\TT a(\theta) x(\alpha-\theta) \dd \theta = \sum_{|n|<m} a_n x_n e^{in\alpha} = \sum_{|n|<m} a_n x_n \cos(n\alpha)$ (by the convolution theorem and the constraint).
Therefore, %
\begin{align*}
    &R[f;\rho](a,b) := \int_{\RR^m} f(\xx)(0)\overline{\rho\left( {\textstyle\sum_{|n|< m}} a_n x_n - b \right)}\dd \xx,\\
    &S[R[f;\rho]](x)(\theta) = \int_{\RR^m\times\RR} R[f;\rho](\aa,b) \sigma\left( {\textstyle\sum_{|n|< m}} a_n x_n \cos(n\theta) - b \right)\dd \aa\dd b = \iiprod{\sigma,\rho} f(x)(\theta).
\end{align*}
\end{ex}

\begin{ex}[Differential operator (with convergence factor)] \label{ex:d.cpconv}
A differential operator $\frac{\dd}{\dd \theta}$ is calculated as $x = \sum_{|n|<m} x_n e^{in\theta} \mapsto f(x) = \sum_{|n|<m} n x_n e^{in\theta}$. Since $\xx \mapsto f(\xx)(0) = \sum_{|n|<m} n x_n$ is not integrable on $\RR^m$, we impose a convergence factor $\phi_t$ as follows.
$f_t(x)(\theta) := f(x)\phi_t(x) = \frac{\dd}{\dd \theta} x(\theta) \phi_t(x) = \sum_{|n|<m} n x_n \phi_t(x) e^{in\theta}$. 
Here, $(\phi_t)_{t > 0} \subset \Sch(\calX)$ is a family of convergence factors that satisfies (1) the first moment $\int_{\calX} |x|_{\calX} |\phi_t(x)| \dd x$ exists at every $t$, (2) $\phi_t \to 1$ in the weak sense as $t \to \infty$, and (3) (continuous and) $(G,T)$-equivariant.
For example, we can take a Gaussian $\phi_t(x) = \exp(-|x|^2_{\calX}/4t)$.
Hence,
    \[R[f_t;\rho](a,b) = \int_{\RR^m} \sum_{|n|<m} n x_n \phi_t(\xx) \overline{\rho\left( {\textstyle\sum_{|n|<m}} a_n x_n - b\right)} \dd \xx.\]
\end{ex}

\subsection{Euclidean group $\gpE(n)$ equivariant map}
\begin{ex}\label{ex:enconv}
The Euclidean group $\gpE(n)$ is a semidirect product $\RR^n \rtimes \gpO(n)$ of the translational group $\RR^n$ and the orthogonal group $\gpO(n)$, which acts on $\RR^n$ as $(U,s)\cdot t := Ut+s$ for any $t \in \RR^n$ and $(U,s) \in O(n) \times \RR^n$. So, put $\calX \subset L^2(\RR^n)$ and $T_{(U,s)}[x](t) := x(U^{-1}(t-s))$. Then,
\begin{align*}
&R[f;\rho](a,b) := \int_{\calX} f(x)(I,0) \overline{\rho\left( \int_{\RR^n} a(t)\overline{x(t)}\dd t - b \right)} \dd x\\
&S[R[f;\rho]](x)(U,t) = \int_{\calX \times \RR} R[f;\rho](a,b) \sigma\left( \int_{\RR^n} a(U^{-1}(t-s))\overline{x(s)} \dd s - b \right) \dd a \dd b = f(x)(U,t)
\end{align*}
\end{ex}
We note that a more memory efficient representation for $L^2(\RR^n)$ and/or a more general representation such as $L^2(\gpSO(2))$ and $L^2(\gpE(3))$, have been developed in the context of \emph{steerable} CNNs \citep{Cohen2017,Weiler2019}.

\section{Discussion}
\subsection{Related Works on (G)CNN Universality} \label{sec:related}%

\paragraph{Non-Group CNN.}
\citet{Zhou2018,Zhou2020a} is the earliest to show the $cc$-universality of deep ReLU (non-group) CNNs.
In \citep{Zhou2020a}, he presented (Theorem~1) the $cc$-universality in $C(\RR^d;\RR)$ in the limit of depth $J \to \infty$, and (Theorem 2) an approximation error rate with respect to $J$.
The CNN is carefully designed so that increasing depth also increases width,
which is not covered in our GCNN. 
\paragraph{Finite Group CNN.}
\citet{Maron2019universality}, \citet{Sannai2019}, \citet{Keriven2019}, \citet{Ravanbakhsh2020} and \citet{Petersen2020}
presented the $cc$-(or $L^p$-)universality results of \emph{finite}-group CNNs. \citet{Maron2019universality} is often cited as one of the earliest publications, where the input space is $\calX=\RR^{n^k \times a}$ ($a$-channel $k$-th order $n$-dimensional tensors), the output space is $\calX’=\RR^{n^l \times b}$ ($b$-channel $l$-th order $n$-dimensional tensors), the group $G$ is a subgroup of a symmetric group $\gpSym_n$, and the group action (or representation) $T$ is the left-translation (or left-regular representation). In this setup, they presented the $cc$-universality of deep-ReLU-GCNNs in the space of continuous $G$-equivariant functions $C_{equi}(\calX;\calX’)$. The proofs are indirect because they are based on  \emph{invariant polynomials} or \emph{MLPs}.
The finite group cases are essentially covered as \refex{deepsets} (Deep Sets).

\paragraph{Lie Group CNN.}
\citet{Yarotsky2021a} carefully designed deep GCNNs with Lie groups acting on \emph{infinite-dimensional} input/output spaces, and show a version of universality in the space of continuous $G$-equivariant functions $C_{equi}( L^2(G;\RR^d) ; L^2(G;\RR^{d’}))$. To be precise, $G$ is either a compact group, translation group $\RR^d$, or 2-dimensional roto-translation group $\gpSE(2)$, and the input/output spaces $\calX$ and $\calX'$ are square-integrable functions on $G$. The proposed networks are not covered in our GCNNs, but several infinite group cases are covered in Examples \ref{ex:cpconv} and \ref{ex:enconv}.

Remarkably, \citet{Kumagai2022} introduced an integral representation that covers LCH groups, and showed the universality. 
The proposed integral representation is based on the Haar measure, thus slightly different from ours. The proofs are indirect because the network is converted to an MLP.

\subsection{Review of Assumptions}

\paragraph{Group $G$.}
We only assume $G$ to be a topological group, to deal with continuous functions on $G$. Thus, a quite large class of groups are covered, for example, all the finite groups such as $\ZZ_n$ and $\gpSym_n$, compact groups such as $\gpSO(n)$ and $\gpU(n)$, and non-compact groups such as $\RR^n$ and $\gpE(n)$ as well. %

\paragraph{Representation Space $\calX$.}
Unlike previous studies, it does \emph{not} need to be a function space such as $C(G)$ and $L^2(G/H)$, but it only needs to be an abstract Hilbert space, which is one of the major advantages for geometric understanding of GCNNs.
On the other hand, we introduce an auxiliary finite-dimensional subspace $\subX$ (and projection $P$), to use the Fourier inversion formula on the finite-dimensional Euclidean space $\RR^m$ in the proof. We conjecture that the extension to an infinite-dimensional setting would be a routine for some specialists in functional analysis. %

\paragraph{Group Representation $T$.} It does \emph{not} need to be unitary, irreducible, nor square-integrable, since the proof is based only on a few basic properties of the linear group representation. %

\paragraph{Network Architecture.} The ridgelet theory supports a wide class of \emph{activation functions}, namely, the tempered distributions ($\Sch'$). The extension to \emph{deep GCNNs} remains an important open question.

\begin{ack}
The authors are grateful to anonymous reviewers for their valuable comments.
This work was supported by JST CREST JPMJCR2015 and JPMJCR1913, JST PRESTO JPMJPR2125, and JST ACT-X JPMJAX2004.

\end{ack}

\bibliographystyle{unsrtnat}
\bibliography{libraryS}

\begin{thebibliography}{67}
\providecommand{\natexlab}[1]{#1}
\providecommand{\url}[1]{\texttt{#1}}
\expandafter\ifx\csname urlstyle\endcsname\relax
  \providecommand{\doi}[1]{doi: #1}\else
  \providecommand{\doi}{doi: \begingroup \urlstyle{rm}\Url}\fi

\bibitem[Bronstein et~al.(2021)Bronstein, Bruna, Cohen, and
  Veli{\v{c}}kovi{\'{c}}]{Bronstein2021}
Michael~M. Bronstein, Joan Bruna, Taco Cohen, and Petar Veli{\v{c}}kovi{\'{c}}.
\newblock \href{http://arxiv.org/abs/2104.13478}{{Geometric Deep Learning:
  Grids, Groups, Graphs, Geodesics, and Gauges}}.
\newblock \emph{arXiv preprint: 2104.13478}, 2021.

\bibitem[Qi et~al.(2017)Qi, Su, Mo, and Guibas]{Qi2017}
Charles~R Qi, Hao Su, Kaichun Mo, and Leonidas~J Guibas.
\newblock
  \href{https://openaccess.thecvf.com/content_cvpr_2017/html/Qi_PointNet_Deep_Learning_CVPR_2017_paper.html}{{PointNet:
  Deep Learning on Point Sets for 3D Classification and Segmentation}}.
\newblock In \emph{Proceedings of the IEEE Conference on Computer Vision and
  Pattern Recognition (CVPR)}, 2017.

\bibitem[Zaheer et~al.(2017)Zaheer, Kottur, Ravanbakhsh, Poczos, Salakhutdinov,
  and Smola]{Zaheer2017}
Manzil Zaheer, Satwik Kottur, Siamak Ravanbakhsh, Barnabas Poczos, Russ~R
  Salakhutdinov, and Alexander~J Smola.
\newblock
  \href{https://papers.nips.cc/paper/2017/hash/f22e4747da1aa27e363d86d40ff442fe-Abstract.html}{{Deep
  Sets}}.
\newblock In \emph{Advances in Neural Information Processing Systems 30}, 2017.

\bibitem[Kondor and Trivedi(2018)]{Kondor2018}
Risi Kondor and Shubhendu Trivedi.
\newblock \href{https://proceedings.mlr.press/v80/kondor18a.html}{{On the
  Generalization of Equivariance and Convolution in Neural Networks to the
  Action of Compact Groups}}.
\newblock In \emph{Proceedings of the 35th International Conference on Machine
  Learning}, volume~80, pages 2747--2755, 2018.

\bibitem[Maron et~al.(2019{\natexlab{a}})Maron, Ben-Hamu, Shamir, and
  Lipman]{Maron2019graph}
Haggai Maron, Heli Ben-Hamu, Nadav Shamir, and Yaron Lipman.
\newblock \href{https://openreview.net/forum?id=Syx72jC9tm}{{Invariant and
  Equivariant Graph Networks}}.
\newblock In \emph{International Conference on Learning Representations},
  2019{\natexlab{a}}.

\bibitem[Cohen et~al.(2018)Cohen, Geiger, K{\"{o}}hler, and Welling]{Cohen2018}
Taco~S Cohen, Mario Geiger, Jonas K{\"{o}}hler, and Max Welling.
\newblock \href{https://openreview.net/forum?id=Hkbd5xZRb}{{Spherical {CNN}s}}.
\newblock In \emph{International Conference on Learning Representations}, 2018.

\bibitem[Cohen et~al.(2019)Cohen, Geiger, and Weiler]{Cohen2019}
Taco~S Cohen, Mario Geiger, and Maurice Weiler.
\newblock
  \href{https://proceedings.neurips.cc/paper/2019/file/b9cfe8b6042cf759dc4c0cccb27a6737-Paper.pdf}{{A
  General Theory of Equivariant CNNs on Homogeneous Spaces}}.
\newblock In \emph{Advances in Neural Information Processing Systems 32}, 2019.

\bibitem[Kondor et~al.(2018)Kondor, Lin, and Trivedi]{Kondor2018a}
Risi Kondor, Zhen Lin, and Shubhendu Trivedi.
\newblock
  \href{https://papers.nips.cc/paper/2018/hash/a3fc981af450752046be179185ebc8b5-Abstract.html}{{Clebsch-Gordan
  Nets: a Fully Fourier Space Spherical Convolutional Neural Network}}.
\newblock In \emph{Advances in Neural Information Processing Systems 31}, 2018.

\bibitem[Zhou(2018)]{Zhou2018}
Ding-Xuan Zhou.
\newblock \href{http://doi.org/10.1142/S0219530518500124}{{Deep distributed
  convolutional neural networks: Universality}}.
\newblock \emph{Analysis and Applications}, 16\penalty0 (06):\penalty0
  895--919, 2018.

\bibitem[Zhou(2020)]{Zhou2020a}
Ding-Xuan Zhou.
\newblock
  \href{http://doi.org/https://doi.org/10.1016/j.acha.2019.06.004}{{Universality
  of deep convolutional neural networks}}.
\newblock \emph{Applied and Computational Harmonic Analysis}, 48\penalty0
  (2):\penalty0 787--794, 2020.

\bibitem[Yarotsky(2021)]{Yarotsky2021a}
Dmitry Yarotsky.
\newblock \href{http://doi.org/10.1007/s00365-021-09546-1}{{Universal
  Approximations of Invariant Maps by Neural Networks}}.
\newblock \emph{Constructive Approximation}, 2021.

\bibitem[Maron et~al.(2019{\natexlab{b}})Maron, Fetaya, Segol, and
  Lipman]{Maron2019universality}
Haggai Maron, Ethan Fetaya, Nimrod Segol, and Yaron Lipman.
\newblock \href{https://proceedings.mlr.press/v97/maron19a.html}{{On the
  Universality of Invariant Networks}}.
\newblock In \emph{Proceedings of the 36th International Conference on Machine
  Learning}, volume~97, pages 4363--4371, 2019{\natexlab{b}}.

\bibitem[Petersen and Voigtlaender(2020)]{Petersen2020}
Philipp Petersen and Felix Voigtlaender.
\newblock \href{http://doi.org/10.1090/proc/14789}{{Equivalence of
  approximation by convolutional neural networks and fully-connected
  networks}}.
\newblock \emph{Proceedings of the American Mathematical Society}, 148\penalty0
  (4):\penalty0 1567--1581, 2020.

\bibitem[Kumagai et~al.(2022)Kumagai, Sannai, and Kawano]{Kumagai2022}
Wataru Kumagai, Akiyoshi Sannai, and Makoto Kawano.
\newblock \href{http://doi.org/10.1080/0952813X.2022.2123563}{{Universal
  approximation with neural networks on function spaces}}.
\newblock \emph{Journal of Experimental \& Theoretical Artificial
  Intelligence}, pages 1--12, 2022.

\bibitem[Okumoto and Suzuki(2022)]{okumoto2022learnability}
Sho Okumoto and Taiji Suzuki.
\newblock \href{https://openreview.net/forum?id=dgxFTxuJ50e}{{Learnability of
  convolutional neural networks for infinite dimensional input via mixed and
  anisotropic smoothness}}.
\newblock In \emph{International Conference on Learning Representations}, 2022.

\bibitem[Cybenko(1989)]{Cybenko1989}
George Cybenko.
\newblock \href{http://doi.org/10.1007/BF02551274}{{Approximation by
  superpositions of a sigmoidal function}}.
\newblock \emph{Mathematics of Control, Signals, and Systems (MCSS)},
  2\penalty0 (4):\penalty0 303--314, 1989.

\bibitem[Hornik et~al.(1989)Hornik, Stinchcombe, and White]{Hornik1989}
Kurt Hornik, Maxwell Stinchcombe, and Halbert White.
\newblock \href{http://doi.org/10.1016/0893-6080(89)90020-8}{{Multilayer
  feedforward networks are universal approximators}}.
\newblock \emph{Neural Networks}, 2\penalty0 (5):\penalty0 359--366, 1989.

\bibitem[Irie and Miyake(1988)]{Irie1988}
Bunpei Irie and Sei Miyake.
\newblock \href{http://doi.org/10.1109/ICNN.1988.23901}{{Capabilities of
  three-layered perceptrons}}.
\newblock In \emph{IEEE International Conference on Neural Networks}, pages
  641--648, 1988.

\bibitem[Funahashi(1989)]{Funahashi1989}
Ken-Ichi Funahashi.
\newblock \href{http://doi.org/10.1016/0893-6080(89)90003-8}{{On the
  approximate realization of continuous mappings by neural networks}}.
\newblock \emph{Neural Networks}, 2\penalty0 (3):\penalty0 183--192, 1989.

\bibitem[Carroll and Dickinson(1989)]{Carroll.Dickinson}
S.~M. Carroll and B.~W. Dickinson.
\newblock \href{http://doi.org/10.1109/IJCNN.1989.118639}{{Construction of
  neural nets using the Radon transform}}.
\newblock In \emph{International Joint Conference on Neural Networks 1989},
  volume~1, pages 607--611. IEEE, 1989.

\bibitem[Ito(1991)]{Ito.Radon}
Yoshifusa Ito.
\newblock \href{http://doi.org/10.1016/0893-6080(91)90075-G}{{Representation of
  functions by superpositions of a step or sigmoid function and their
  applications to neural network theory}}.
\newblock \emph{Neural Networks}, 4\penalty0 (3):\penalty0 385--394, 1991.

\bibitem[Mhaskar and Micchelli(1992)]{Mhaskar1992}
H.N Mhaskar and Charles~A Micchelli.
\newblock \href{http://doi.org/10.1016/0196-8858(92)90016-P}{{Approximation by
  superposition of sigmoidal and radial basis functions}}.
\newblock \emph{Advances in Applied Mathematics}, 13\penalty0 (3):\penalty0
  350--373, 1992.

\bibitem[Mhaskar(1996)]{Mhaskar1996}
H.~N. Mhaskar.
\newblock \href{http://doi.org/10.1162/neco.1996.8.1.164}{{Neural Networks for
  Optimal Approximation of Smooth and Analytic Functions}}.
\newblock \emph{Neural Computation}, 8:\penalty0 164--177, 1996.

\bibitem[Barron(1993)]{Barron1993}
Andrew~R Barron.
\newblock \href{http://doi.org/10.1109/18.256500}{{Universal approximation
  bounds for superpositions of a sigmoidal function}}.
\newblock \emph{IEEE Transactions on Information Theory}, 39\penalty0
  (3):\penalty0 930--945, 1993.

\bibitem[Leshno et~al.(1993)Leshno, Lin, Pinkus, and Schocken]{Leshno1993}
Moshe Leshno, Vladimir~Ya. Lin, Allan Pinkus, and Shimon Schocken.
\newblock \href{http://doi.org/10.1016/S0893-6080(05)80131-5}{{Multilayer
  feedforward networks with a nonpolynomial activation function can approximate
  any function}}.
\newblock \emph{Neural Networks}, 6\penalty0 (6):\penalty0 861--867, 1993.

\bibitem[Murata(1996)]{Murata1996}
Noboru Murata.
\newblock \href{http://doi.org/10.1016/0893-6080(96)00000-7}{{An integral
  representation of functions using three-layered betworks and their
  approximation bounds}}.
\newblock \emph{Neural Networks}, 9\penalty0 (6):\penalty0 947--956, 1996.

\bibitem[Cand{\`{e}}s(1998)]{Candes.PhD}
Emmanuel~Jean Cand{\`{e}}s.
\newblock
  \emph{\href{https://searchworks.stanford.edu/view/9949708}{{Ridgelets: theory
  and applications}}}.
\newblock PhD thesis, Standford University, 1998.

\bibitem[Rubin(1998)]{Rubin.calderon}
Boris Rubin.
\newblock \href{http://doi.org/10.1007/BF02475988}{{The Calder{\'{o}}n
  reproducing formula, windowed X-ray transforms, and Radon transforms in
  $L^p$-spaces}}.
\newblock \emph{Journal of Fourier Analysis and Applications}, 4\penalty0
  (2):\penalty0 175--197, 1998.

\bibitem[Telgarsky(2016)]{Telgarsky2016}
Matus Telgarsky.
\newblock \href{https://proceedings.mlr.press/v49/telgarsky16.html}{{Benefits
  of depth in neural networks}}.
\newblock In \emph{29th Annual Conference on Learning Theory}, pages 1--23,
  2016.

\bibitem[Yarotsky(2017)]{Yarotsky2017}
Dmitry Yarotsky.
\newblock \href{http://doi.org/10.1016/j.neunet.2017.07.002}{{Error bounds for
  approximations with deep ReLU networks}}.
\newblock \emph{Neural Networks}, 94:\penalty0 103--114, 2017.

\bibitem[Petersen and Voigtlaender(2018)]{Petersen2018}
Philipp Petersen and Felix Voigtlaender.
\newblock
  \href{http://doi.org/https://doi.org/10.1016/j.neunet.2018.08.019}{{Optimal
  approximation of piecewise smooth functions using deep ReLU neural
  networks}}.
\newblock \emph{Neural Networks}, 108:\penalty0 296--330, 2018.

\bibitem[Cohen and Welling(2017)]{Cohen2017}
Taco~S Cohen and Max Welling.
\newblock \href{https://openreview.net/forum?id=rJQKYt5ll}{{Steerable CNNs}}.
\newblock In \emph{International Conference on Learning Representations 2017},
  pages 1--14, 2017.

\bibitem[Jacot et~al.(2018)Jacot, Gabriel, and Hongler]{Jacot2018}
Arthur Jacot, Franck Gabriel, and Clement Hongler.
\newblock
  \href{http://papers.nips.cc/paper/8076-neural-tangent-kernel-convergence-and-generalization-in-neural-networks.pdf}{{Neural
  Tangent Kernel: Convergence and Generalization in Neural Networks}}.
\newblock In \emph{Advances in Neural Information Processing Systems 31}, pages
  8571--8580, 2018.

\bibitem[Lee et~al.(2019)Lee, Xiao, Schoenholz, Bahri, Sohl-Dickstein, and
  Pennington]{Lee2019}
Jaehoon Lee, Lechao Xiao, Samuel~S. Schoenholz, Yasaman Bahri, Jascha
  Sohl-Dickstein, and Jeffrey Pennington.
\newblock
  \href{http://papers.nips.cc/paper/9063-wide-neural-networks-of-any-depth-evolve-as-linear-models-under-gradient-descent/}{{Wide
  Neural Networks of Any Depth Evolve as Linear Models Under Gradient
  Descent}}.
\newblock In \emph{Advances in Neural Information Processing Systems 32}, pages
  8572--8583, 2019.

\bibitem[Arora et~al.(2019)Arora, Du, Hu, Li, and Wang]{Arora2019b}
Sanjeev Arora, Simon~S. Du, Wei Hu, Zhiyuan Li, and Ruosong Wang.
\newblock \href{http://proceedings.mlr.press/v97/arora19a.html}{{Fine-Grained
  Analysis of Optimization and Generalization for Overparameterized Two-Layer
  Neural Networks}}.
\newblock In \emph{Proceedings of the 36th International Conference on Machine
  Learning}, volume~97, pages 322--332, 2019.

\bibitem[Chizat et~al.(2019)Chizat, Oyallon, and Bach]{Chizat2019}
Lénaïc Chizat, Edouard Oyallon, and Francis Bach.
\newblock
  \href{http://papers.nips.cc/paper/8559-on-lazy-training-in-differentiable-programming.pdf}{{On
  Lazy Training in Differentiable Programming}}.
\newblock In \emph{Advances in Neural Information Processing Systems 32}, pages
  2937--2947, 2019.

\bibitem[Frankle and Carbin(2019)]{Frankle2019}
Jonathan Frankle and Michael Carbin.
\newblock \href{https://openreview.net/forum?id=rJl-b3RcF7}{{The Lottery Ticket
  Hypothesis: Finding Sparse, Trainable Neural Networks}}.
\newblock In \emph{International Conference on Learning Representations}, pages
  1--42, 2019.

\bibitem[Nitanda and Suzuki(2017)]{Nitanda2017}
Atsushi Nitanda and Taiji Suzuki.
\newblock \href{http://arxiv.org/abs/1712.05438}{{Stochastic Particle Gradient
  Descent for Infinite Ensembles}}.
\newblock \emph{arXiv preprint: 1712.05438}, 2017.

\bibitem[Mei et~al.(2018)Mei, Montanari, and Nguyen]{Mei2018}
Song Mei, Andrea Montanari, and Phan-Minh Nguyen.
\newblock \href{http://doi.org/10.1073/PNAS.1806579115}{{A mean field view of
  the landscape of two-layer neural networks}}.
\newblock \emph{Proceedings of the National Academy of Sciences}, 115\penalty0
  (33):\penalty0 E7665--E7671, 2018.

\bibitem[Rotskoff and Vanden-Eijnden(2018)]{Rotskoff2018}
Grant Rotskoff and Eric Vanden-Eijnden.
\newblock
  \href{http://papers.nips.cc/paper/7945-parameters-as-interacting-particles-long-time-convergence-and-asymptotic-error-scaling-of-neural-networks.pdf}{{Parameters
  as interacting particles: long time convergence and asymptotic error scaling
  of neural networks}}.
\newblock In \emph{Advances in Neural Information Processing Systems 31}, pages
  7146--7155, 2018.

\bibitem[Chizat and Bach(2018)]{Chizat2018}
Lénaïc Chizat and Francis Bach.
\newblock
  \href{https://papers.nips.cc/paper/7567-on-the-global-convergence-of-gradient-descent-for-over-parameterized-models-using-optimal-transport/}{{On
  the Global Convergence of Gradient Descent for Over-parameterized Models
  using Optimal Transport}}.
\newblock In \emph{Advances in Neural Information Processing Systems 32}, pages
  3036--3046, 2018.

\bibitem[Sirignano and Spiliopoulos(2020)]{Sirignano2020}
Justin Sirignano and Konstantinos Spiliopoulos.
\newblock \href{http://doi.org/10.1137/18M1192184}{{Mean Field Analysis of
  Neural Networks: A Law of Large Numbers}}.
\newblock \emph{SIAM Journal on Applied Mathematics}, 80\penalty0 (2):\penalty0
  725--752, 2020.

\bibitem[Suzuki(2020)]{Suzuki2020}
Taiji Suzuki.
\newblock
  \href{https://proceedings.neurips.cc/paper/2020/hash/df1a336b7e0b0cb186de6e66800c43a9-Abstract.html}{{Generalization
  bound of globally optimal non-convex neural network training: Transportation
  map estimation by infinite dimensional Langevin dynamics}}.
\newblock In \emph{Advances in Neural Information Processing Systems 33}, pages
  19224--19237, 2020.

\bibitem[Sonoda et~al.(2021{\natexlab{a}})Sonoda, Ishikawa, and
  Ikeda]{Sonoda2021ghost}
Sho Sonoda, Isao Ishikawa, and Masahiro Ikeda.
\newblock \href{http://arxiv.org/abs/2106.04770}{{Ghosts in Neural Networks:
  Existence, Structure and Role of Infinite-Dimensional Null Space}}.
\newblock \emph{arXiv preprint: 2106.04770}, 2021{\natexlab{a}}.

\bibitem[Sonoda et~al.(2021{\natexlab{b}})Sonoda, Ishikawa, and
  Ikeda]{Sonoda2021aistats}
Sho Sonoda, Isao Ishikawa, and Masahiro Ikeda.
\newblock \href{http://proceedings.mlr.press/v130/sonoda21a.html}{{Ridge
  Regression with Over-Parametrized Two-Layer Networks Converge to Ridgelet
  Spectrum}}.
\newblock In \emph{Proceedings of The 24th International Conference on
  Artificial Intelligence and Statistics (AISTATS) 2021}, volume 130, pages
  2674--2682, 2021{\natexlab{b}}.

\bibitem[Savarese et~al.(2019)Savarese, Evron, Soudry, and
  Srebro]{Savarese2019}
Pedro Savarese, Itay Evron, Daniel Soudry, and Nathan Srebro.
\newblock \href{http://proceedings.mlr.press/v99/savarese19a.html}{{How do
  infinite width bounded norm networks look in function space?}}
\newblock In \emph{Proceedings of the 32nd Conference on Learning Theory},
  volume~99, pages 2667--2690, 2019.

\bibitem[Ongie et~al.(2020)Ongie, Willett, Soudry, and Srebro]{Ongie2020}
Greg Ongie, Rebecca Willett, Daniel Soudry, and Nathan Srebro.
\newblock \href{https://openreview.net/forum?id=H1lNPxHKDH}{{A Function Space
  View of Bounded Norm Infinite Width ReLU Nets: The Multivariate Case}}.
\newblock In \emph{International Conference on Learning Representations}, 2020.

\bibitem[Parhi and Nowak(2021)]{Parhi2021}
Rahul Parhi and Robert~D Nowak.
\newblock \href{http://jmlr.org/papers/v22/20-583.html}{{Banach Space
  Representer Theorems for Neural Networks and Ridge Splines}}.
\newblock \emph{Journal of Machine Learning Research}, 22\penalty0
  (43):\penalty0 1--40, 2021.

\bibitem[Unser(2019)]{Unser2019}
Michael Unser.
\newblock \href{http://jmlr.org/papers/v20/18-418.html}{{A Representer Theorem
  for Deep Neural Networks}}.
\newblock \emph{Journal of Machine Learning Research}, 20\penalty0
  (110):\penalty0 1--30, 2019.

\bibitem[Donoho(2002)]{Donoho2002}
David~L Donoho.
\newblock \href{http://arxiv.org/abs/math/0212395}{{Emerging applications of
  geometric multiscale analysis}}.
\newblock \emph{Proceedings of the ICM, Beijing 2002}, I:\penalty0 209--233,
  2002.

\bibitem[Starck et~al.(2010)Starck, Murtagh, and Fadili]{Starck2010}
Jean-Luc Starck, Fionn Murtagh, and Jalal~M. Fadili.
\newblock \href{http://doi.org/10.1017/CBO9780511730344.006}{{The ridgelet and
  curvelet transforms}}.
\newblock In \emph{Sparse Image and Signal Processing: Wavelets, Curvelets,
  Morphological Diversity}, pages 89--118. Cambridge University Press, 2010.

\bibitem[Kutyniok and Labate(2012)]{Kutyniok2012}
Gitta Kutyniok and Demetrio Labate.
\newblock \emph{\href{http://doi.org/10.1007/978-0-8176-8316-0}{{Shearlets:
  Multiscale Analysis for Multivariate Data}}}.
\newblock Applied and Numerical Harmonic Analysis. Birkh{\"{a}}user Boston, 1
  edition, 2012.

\bibitem[Sonoda and Murata(2017)]{Sonoda2015acha}
Sho Sonoda and Noboru Murata.
\newblock \href{http://doi.org/10.1016/j.acha.2015.12.005}{{Neural network with
  unbounded activation functions is universal approximator}}.
\newblock \emph{Applied and Computational Harmonic Analysis}, 43\penalty0
  (2):\penalty0 233--268, 2017.

\bibitem[Kostadinova et~al.(2014)Kostadinova, Pilipovi{\'{c}}, Saneva, and
  Vindas]{Kostadinova2014}
S~Kostadinova, S~Pilipovi{\'{c}}, K~Saneva, and J~Vindas.
\newblock \href{http://doi.org/10.1080/10652469.2013.853057}{{The ridgelet
  transform of distributions}}.
\newblock \emph{Integral Transforms and Special Functions}, 25\penalty0
  (5):\penalty0 344--358, 2014.

\bibitem[Sonoda et~al.(2022)Sonoda, Ishikawa, and Ikeda]{sonoda2022symmetric}
Sho Sonoda, Isao Ishikawa, and Masahiro Ikeda.
\newblock
  \href{https://proceedings.mlr.press/v162/sonoda22a.html}{{Fully-Connected
  Network on Noncompact Symmetric Space and Ridgelet Transform based on
  Helgason-Fourier Analysis}}.
\newblock In \emph{Proceedings of the 39th International Conference on Machine
  Learning}, volume 162, pages 20405--20422, 2022.

\bibitem[Grafakos(2008)]{Grafakos.classic}
Loukas Grafakos.
\newblock \emph{\href{http://doi.org/10.1007/978-0-387-09432-8}{{Classical
  Fourier Analysis}}}.
\newblock Graduate Texts in Mathematics. Springer New York, second edition,
  2008.

\bibitem[Gel'fand and Shilov(1964)]{GelfandShilov}
I.~M. Gel'fand and G.~E. Shilov.
\newblock \emph{{Generalized Functions, Vol. 1: Properties and Operations}}.
\newblock Academic Press, New York, 1964.

\bibitem[Folland(2015)]{Folland2015}
Gerald~B. Folland.
\newblock \emph{\href{https://doi.org/10.1201/b19172}{A Course in Abstract
  Harmonic Analysis}}.
\newblock Chapman and Hall/CRC, New York, second edition, 2015.

\bibitem[Sriperumbudur et~al.(2010)Sriperumbudur, Fukumizu, and
  Lanckriet]{Sriperumbudur2010}
Bharath~K. Sriperumbudur, Kenji Fukumizu, and Gert R.~G. Lanckriet.
\newblock
  \href{https://jmlr.org/papers/v12/sriperumbudur11a.html}{{Universality,
  Characteristic Kernels and RKHS Embedding of Measures}}.
\newblock \emph{Journal of Machine Learning Research}, 12\penalty0
  (Jul):\penalty0 2389--2410, 2010.

\bibitem[Pinkus(1999)]{Pinkus.survey}
Allan Pinkus.
\newblock \href{http://doi.org/10.1017/S0962492900002919}{{Approximation theory
  of the MLP model in neural networks}}.
\newblock \emph{Acta Numerica}, 8:\penalty0 143--195, 1999.

\bibitem[Krizhevsky et~al.(2012)Krizhevsky, Sutskever, and
  Hinton]{Krizhevsky2012}
Alex Krizhevsky, Ilya Sutskever, and Geoffrey~E. Hinton.
\newblock
  \href{http://papers.nips.cc/paper/4824-imagenet-classification-with-deep-convolutional-neural-networks.pdf}{{ImageNet
  Classification with Deep Convolutional Neural Networks}}.
\newblock In \emph{Advances in Neural Information Processing Systems 25}, pages
  1097--1105, 2012.

\bibitem[Zeiler and Fergus(2014)]{Zeiler2014}
Matthew~D. Zeiler and Rob Fergus.
\newblock \href{https://doi.org/10.1007/978-3-319-10590-1_53}{{Visualizing and
  Understanding Convolutional Networks}}.
\newblock In \emph{European Conference on Computer Vision}, pages 818--833,
  2014.

\bibitem[Weiler and Cesa(2019)]{Weiler2019}
Maurice Weiler and Gabriele Cesa.
\newblock
  \href{https://papers.nips.cc/paper/2019/hash/45d6637b718d0f24a237069fe41b0db4-Abstract.html}{{General
  E(2)-Equivariant Steerable CNNs}}.
\newblock In \emph{Advances in Neural Information Processing Systems 32}, 2019.

\bibitem[Sannai et~al.(2019)Sannai, Takai, and Cordonnier]{Sannai2019}
Akiyoshi Sannai, Yuuki Takai, and Matthieu Cordonnier.
\newblock \href{http://doi.org/10.48550/arxiv.1903.01939}{{Universal
  approximations of permutation invariant/equivariant functions by deep neural
  networks}}.
\newblock \emph{arXiv preprint: 1903.01939}, 2019.

\bibitem[Keriven and Peyr{\'{e}}(2019)]{Keriven2019}
Nicolas Keriven and Gabriel Peyr{\'{e}}.
\newblock
  \href{https://proceedings.neurips.cc/paper/2019/hash/ea9268cb43f55d1d12380fb6ea5bf572-Abstract.html}{{Universal
  Invariant and Equivariant Graph Neural Networks}}.
\newblock In \emph{Advances in Neural Information Processing Systems 32}, 2019.

\bibitem[Ravanbakhsh(2020)]{Ravanbakhsh2020}
Siamak Ravanbakhsh.
\newblock
  \href{https://proceedings.mlr.press/v119/ravanbakhsh20a.html}{{Universal
  Equivariant Multilayer Perceptrons}}.
\newblock In \emph{Proceedings of the 37th International Conference on Machine
  Learning}, volume 119 of \emph{Proceedings of Machine Learning Research},
  pages 7996--8006, 2020.

\bibitem[Kainen et~al.(2013)Kainen, K\r{u}rkov\'{a}, and
  Sanguineti]{kainen.survey}
P~C Kainen, V\u{e}ra K\r{u}rkov\'{a}, and Marcello Sanguineti.
\newblock \href{http://doi.org/10.1007/978-3-642-36657-4}{{Approximating
  multivariable functions by feedforward neural nets}}.
\newblock In \emph{Handbook on Neural Information Processing}, volume~49 of
  \emph{Intelligent Systems Reference Library}, pages 143--181. Springer Berlin
  Heidelberg, 2013.

\end{thebibliography}

\newpage
\section*{Checklist}

\begin{enumerate}

\item For all authors...
\begin{enumerate}
  \item Do the main claims made in the abstract and introduction accurately reflect the paper's contributions and scope?
    \answerYes{}
  \item Did you describe the limitations of your work?
    \answerYes{}
  \item Did you discuss any potential negative societal impacts of your work?
    \answerNo{}
  \item Have you read the ethics review guidelines and ensured that your paper conforms to them?
    \answerYes{}
\end{enumerate}

\item If you are including theoretical results...
\begin{enumerate}
  \item Did you state the full set of assumptions of all theoretical results?
    \answerYes{}
        \item Did you include complete proofs of all theoretical results?
    \answerYes{}
\end{enumerate}

\item If you ran experiments...
\begin{enumerate}
  \item Did you include the code, data, and instructions needed to reproduce the main experimental results (either in the supplemental material or as a URL)?
    \answerNA{}
  \item Did you specify all the training details (e.g., data splits, hyperparameters, how they were chosen)?
    \answerNA{}
        \item Did you report error bars (e.g., with respect to the random seed after running experiments multiple times)?
    \answerNA{}
        \item Did you include the total amount of compute and the type of resources used (e.g., type of GPUs, internal cluster, or cloud provider)?
    \answerNA{}
\end{enumerate}

\item If you are using existing assets (e.g., code, data, models) or curating/releasing new assets...
\begin{enumerate}
  \item If your work uses existing assets, did you cite the creators?
    \answerNA{}
  \item Did you mention the license of the assets?
    \answerNA{}
  \item Did you include any new assets either in the supplemental material or as a URL?
    \answerNA{}
  \item Did you discuss whether and how consent was obtained from people whose data you're using/curating?
    \answerNA{}
  \item Did you discuss whether the data you are using/curating contains personally identifiable information or offensive content?
    \answerNA{}
\end{enumerate}

\item If you used crowdsourcing or conducted research with human subjects...
\begin{enumerate}
  \item Did you include the full text of instructions given to participants and screenshots, if applicable?
    \answerNA{}
  \item Did you describe any potential participant risks, with links to Institutional Review Board (IRB) approvals, if applicable?
    \answerNA{}
  \item Did you include the estimated hourly wage paid to participants and the total amount spent on participant compensation?
    \answerNA{}
\end{enumerate}

\end{enumerate}

\newpage
\appendix
\section{Proofs}
\paragraph{Additional Notation} 
In the proofs, we use two symbols $\widehat{\cdot}$ and $\cdot^\sharp$ for the Fourier transforms in $x \in \calX$ and $b \in \KK$, respectively. For example,
\begin{align*}
    &\widehat{f}(\xi) := \int_{\calX}f(x)e^{-i \iprod{\xi,x}_{\calX}} \dd x, \quad \xi \in \calX\\
    &\rho^\sharp(\omega) := \int_\KK \rho(b)e^{-i\omega b}\dd b, \quad \omega \in \KK\\
    &\gamma^\sharp(a,\omega) = \int_\KK \gamma(a,b)e^{-i\omega b}\dd b, \quad (a,\omega) \in \calX\times\KK.
\end{align*}
With a slight abuse of notation, when $\sigma$ is a tempered distribution (i.e., $\sigma \in \Sch'(\KK)$), then $\sigma^\sharp$ is understood as the Fourier transform of distributions. Namely, $\sigma^\sharp$ is another tempered distribution satisfying $\int_\KK \sigma^\sharp(\omega)\phi(\omega)\dd\omega = \int_\KK \sigma(\omega) \phi^\sharp(\omega)\dd\omega$ for any test function $\phi \in \Sch(\KK)$.

For any integer $d >0$ and vector $ \vv \in \RR^d$, $|\vv|$ denotes the Euclidean norm, and $\iprod{\vv} := \sqrt{1 + |\vv|^2}$. For any positive number $t > 0$, $\triangle^{t/2}$ and $\iprod{\triangle}^t$ denote fractional differential operators defined as Fourier multipliers: for any $\phi \in \Sch'(\RR^d)$,
\begin{align}
&\triangle^{t/2}[\phi](\vv) := \frac{1}{(2\pi)^d} \int_{\RR^d} |\uu|^t \widehat{\phi}(\uu) e^{i\uu\cdot\vv}\dd\uu,\\
&\iprod{\triangle}^{t/2}[\phi](\vv) := \frac{1}{(2\pi)^d} \int_{\RR^d} (1+|\uu|^2)^{t/2} \widehat{\phi}(\uu) e^{i\uu\cdot\vv}\dd\uu.
\end{align}
In particular when $t=2$, $\triangle^{t/2}$ coincides with the ordinary Laplacian on $\RR^d$. 

\newcommand{\barH}{\calX}
\subsection{\refthm{gconv.fdim.elem.equiv}} \label{sec:proof.reconst}
\begin{proof}

    In the following, we fix a representation $T:G\to GL(\barH)$ of a group $G$ on a (potentially infinite-dimensional) Hilbert space $\barH$ over a field $\KK$ equipped with inner product $\iprod{\cdot,\cdot}_{\barH}$, which is a $G$-invariant vector space: $T_G[\barH] = \barH$, and a finite-dimensional closed subspace $\subX \subset \barH$ equipped with the Lebesgue measure $\lambda$.
    Let $\perpX$ be the orthogonal complement of $\subX$ in $\barH$, i.e., $x \oplus y \in \subX \oplus \perpX = \barH$, and 
    let $P:\barH\to\subX$ denote the orthogonal projection onto $\subX$.
    Let $m := \dim_\RR \subX$ denotes the real dimension of $\subX$, and let $k := \dim_\RR\KK$ denotes the real dimension of $\KK$, which is either $1$ or $2$.

    Without loss of generality, we can assume
        (A') that $\gamma(a,\bullet) * \sigma \in \Sch(\KK)$ for a.e. $a \in \subX$, and 
        (A'') that $\gamma^\sharp \sigma^\sharp \in L^1(\subX\times\KK)$,
    which will be eventually justified because later in \refeq{sov}, we set $\gamma^\sharp(a,\omega) = \widehat{f}(\omega a)\overline{\rho^\sharp(\omega)}$.

    \paragraph{Step~1 (Fourier expression).}
    Using an identity: For any function $\phi \in \Sch(\KK)$ and $b \in \KK$, $\phi(b) = \frac{1}{(2\pi)^k}\int_\KK \phi^\sharp(\omega) e^{ib\omega}\dd\omega$, namely the Fourier inversion formula, we can turn $S$ into a \emph{Fourier expression}:
    \begin{align}
        S[\gamma](x)(g) 
        &= \frac{1}{(2\pi)^k}\int_{\subX} \int_{\KK} \gamma(a,b) \sigma(\iprod{T_{g^{-1}}[x],a}_{\calX}-b) \dd b \dd a \\
        &= \frac{1}{(2\pi)^k}\int_{\subX} \int_{\KK} \gamma^\sharp(a,\omega) \sigma^\sharp(\omega)\exp\left( i \omega \iprod{T_{g^{-1}}[x],a}_{\calX}\right) \dd \omega \dd a.
    \end{align}
    By the assumption (A'), the first equation holds at every point $b = ( a *_T x )(g)$, and by the assumption (A''), the Fourier expression is \emph{uniformly} absolutely convergent:
    \begin{align}
        \int_{\subX\times\KK}| \gamma^\sharp(a,\omega) \sigma^\sharp(\omega) \exp( i\omega (a * x)(g) ) | \dd a \dd \omega 
        = \| \gamma^\sharp \sigma^\sharp \|_{L^1(\subX \times \KK)} < \infty,
    \end{align}
    for \emph{all} $(x,g) \in \calX \times G$.
    Hence, we can change the order of integration freely.
    
    \paragraph{Step~2 (Reconstruction).}
    By changing the variables as $(a,\omega) = (\xi/\omega,\omega)$ with $\dd a \dd \omega = |\omega|^{-m} \dd \xi \dd \omega$, we have
    \begin{align}
        S[\gamma](x)(g) = \frac{1}{(2\pi)^k}\int_{\subX \times \KK} \gamma^\sharp(\xi/\omega,\omega) \sigma^\sharp(\omega)\exp\left( i \iprod{T_{g^{-1}}[x],\xi}_{\calX}\right) |\omega|^{-m} \dd \omega \dd \xi.
    \end{align}
    Hence, using a given $f$ satisfying the assumptions (A1) and (A2), and some function $\rho \in \Sch(\KK)$, suppose that $\gamma_{f,\rho}$ satisfies the following \emph{separation-of-variables} form:
    \begin{align}
        \gamma_{f,\rho}^\sharp(\xi/\omega,\omega) = \widehat{f}(\xi)(e)\overline{\rho^\sharp(\omega)}. \label{eq:sov}
    \end{align}
    Then, 
    \begin{align}
        S[\gamma_{f,\rho}](x)(g) 
        &=
        \frac{1}{(2\pi)^k}\int_{\subX \times \KK} \widehat{f}(\xi)(e)\overline{\rho^\sharp(\omega)} \sigma^\sharp(\omega)\exp\left( i \iprod{T_{g^{-1}}[x],\xi}_{\calX}\right) |\omega|^{-m} \dd \omega \dd \xi\\
        &=\left( (2\pi)^{m-k} \int_{\KK} \sigma^\sharp(\omega) \overline{\rho^\sharp(\omega)}|\omega|^{-m}\dd\omega \right) \notag\\
        &\qquad \times \left( \frac{1}{(2\pi)^m}\int_{\subX} \widehat{f}(\xi)(e)\exp\left( i \iprod{T_{g^{-1}}[x],\xi}_{\calX}\right) \dd \xi\right)\\
        &\stareq \iiprod{\sigma,\rho} f(P T_{g^{-1}}[x])(e)\\
        &=\iiprod{\sigma,\rho}f(x)(g).
    \end{align}
    where we put 
    \begin{align}
        \iiprod{\sigma,\rho} := (2\pi)^{m-k} \int_{\KK} \sigma^\sharp(\omega) \overline{\rho^\sharp(\omega)}|\omega|^{-m}\dd\omega.
    \end{align}
    Here, the equality $\stareq$ holds at every continuous point $x_c$ of $f$ for (A2a), in $L^2$ for (A2b), and  in $\Sch'$ for (A2c), respectively.

    \paragraph{Step~3 (Ridgelet transform).}
    Since we put
    \begin{align}
        \gamma_{f,\rho}^\sharp(a,\omega) = \widehat{f}(\omega a)(e)\overline{\rho^\sharp(\omega)},
    \end{align}
    it is calculated as
    \begin{align}
        \gamma_{f,\rho}(a,b) 
        &= \frac{1}{(2\pi)^k} \int_{\KK} \widehat{f}(\omega a)(e)\overline{\rho^\sharp(\omega)} e^{i\omega b}\dd \omega\\
        &= \frac{1}{(2\pi)^k} \int_{\KK \times \subX} f(x)(e) \overline{\rho^\sharp(\omega)} e^{i\omega (b-\iprod{ a,x}_{\calX})}\dd \omega\dd x\\
        &= \int_{\subX} f(x)(e) \overline{\rho( \iprod{a,x}_{\calX} - b )}\dd x,
    \end{align}
    which is the definition of the ridgelet transform for GCNN.
\end{proof}

\subsection{\refthm{gcnn.universality}} \label{sec:proof.universality}

\begin{proof}
Fix arbitrary compact sets $K \subset \subX$ and $L \subset G$,
positive number $\eps>0$,
and function $f \in C_{equi}(K; C(G))$. 
An $n$-term finite $(G,T)$-CNN is given by
\begin{align}
    f_n(x)(g) := \sum_{i=1}^n c_i \sigma\left( (a_i *_T x)(g) - b_i \right), \quad x \in \subX, \ g \in G
\end{align}
with parameters $(a_i,b_i,c_i) \in \subX \times \KK \times \CC$.
Observe that any finite $(G,T)$-CNN is $(G,T)$-equivariant, that is,
\begin{align}
    f_n(T_g[x])(h)
    &= \sum_{i=1}^n c_i \sigma\left( \iprod{T_{(g^{-1}h)^{-1}} x,a_i}_{\calX} - b_i \right)
    = f_n(x)(g^{-1}h).
\end{align}
Put $\overline{K} := \{ T_{g^{-1}}[x] \mid x \in K, g \in L\}$, which is compact because $T$ is continuous, and 
put $f_e(x) := f(x)(e)$, which is compactly supported, i.e., $f_e \in C(K) \subset C(\overline{K})$.
By \refthm{cc}, there exist a finite number $N \in \NN$ and an $N$-term $\CC$-valued fully-connected network $F_N(x) = \sum_{i=1}^N c_i \sigma( \iprod{a_i,x}_{\calX} - b_i )$ satisfying $\|F_N - f_e \|_{C(\overline{K})} < \eps$. Put $f_N(x)(g) := F_N(T_{g^{-1}}[x])$. Then, it is a $(G,T)$-CNN because
\begin{align}
    f_N(x)(g)
    = \sum_{i=1}^N c_i \sigma\left( \iprod{T_{g^{-1}},a_i}_{\calX} - b_i \right)
    = \sum_{i=1}^N c_i \sigma\left( (a_i *_T x)(g) - b_i \right),
\end{align}
and it is an $\eps$-neighbour of $f$ because
\begin{align}
    \| f_N - f \|_{C( K; C(L) )}
    &= \sup_{x \in K} \sup_{g \in L} | f_N(x)(g) - f(x)(g) |\\
    &= \sup_{x \in K} \sup_{g \in L} | F_N( T_{g^{-1}}[x]) - f_e(T_{g^{-1}}[x]) |\\
    &= \sup_{g \in L} \sup_{x' \in \overline{K}} | F_N(x') - f_e(x') |, \quad x' = T_{g^{-1}}[x]\\
    &< \eps,
\end{align}
which concludes the assertion.
\end{proof}

\begin{thm}[$cc$-universality of scalar-valued finite fully-connected NNs on $\RR^m$] \label{thm:cc}
Suppose that 
\begin{enumerate}
    \item $\calX = \subX = \RR^m$,
    \item $f \in C(\calX;\CC)$ (not vector-valued $C(\calX;\CC^G)$ but scalar-valued), and 
    \item there exists $k \ge 0$ and $\theta>0$ such that 
    $\Delta_\theta^k[\sigma] \in L^\infty(\RR)$ and Lipschitz continuous.
\end{enumerate}
    Then, the finite neural networks of the form $f_n(\xx) = \sum_{i=1}^n c_i \sigma(\aa_i\cdot\xx-b_i)$ are $cc$-universal, that is, for any compact set $K \subset \RR^m$, positive number $\eps>0$, and continuous function $f \in C(K)$, there exists a finite network $f_n$ such that $\| f - f_n \|_{C(K)} < \eps$.
\end{thm}

\newcommand{\fa}{{f_c}}
\newcommand{\fb}{{f_\cube}}
\newcommand{\fc}{{f_n}}
\begin{proof} 
Since $\sum_{i=1}^n c_i \Delta_\theta^k[\sigma](\aa_i\cdot\xx-b_i)$ is rewritten as another finite model $\sum_{i=1}^{n'} c_i' \sigma(\aa_i'\cdot\xx-b_i')$, it suffice to consider the case $k=0$. In the following, we assume that $\sigma (= \Delta_\theta^0[\sigma])$ is bounded and Lipschitz continuous. 
\paragraph{Step~1 ($f \sim \fa$).}
By the density of $C_c^\infty(\RR^m)$ in $C(K)$ with respect to the uniform norm, we can take a compactly-supported smooth function $\fa \in C_c^\infty(\RR^m)$ satisfying $\| f - \fa \|_{C(K)} < \eps/3$. Since $\fa$ is sufficiently smooth and integrable, there exists a compactly-supported smooth function $\rho \in C_c^\infty(\RR)$ such that 
\begin{align}
    S[R[\fa;\rho]](\xx) = \fa(\xx) \mbox{ at every point } \xx \in \RR^m.
\end{align}
For example, 
take a compactly-supported smooth function $\rho_0 \in C_c^\infty(\KK)$, write $k = \dim_{\RR}\KK (= 1 \mbox{ or }2)$, and put $\rho(b) := \triangle_b^{m/2}[\rho_0](b) = (2\pi)^{-k}\int_\KK |\omega|^m \rho_0^\sharp(\omega) e^{ib\cdot\omega}\dd\omega$. Then, $\iiprod{\sigma,\rho} = (2\pi)^{m-k}\int_{\KK} \sigma^\sharp(\omega) \overline{\rho^\sharp(\omega)}|\omega|^{-m}\dd \omega = (2\pi)^{m-k}\int_{\KK} \sigma^\sharp(\omega) \overline{\rho_0^\sharp(\omega)} \dd \omega = (2\pi)^m \int_{\KK} \sigma(b) \overline{\rho_0(b)} \dd b = \iprod{\sigma,\rho_0}_{L^2(\KK)}$, which is an ordinary functional inner product, and it is easy to find a $\rho_0$ satisfying $\iprod{\sigma,\rho_0}_{L^2(\KK)} \neq 0$. By normalizing $\rho' := \rho/\iiprod{\sigma,\rho}$, we can find the $\rho'$. We refer to \citet{Sonoda2015acha} and \citet{Sonoda2021ghost} for more details on the scalar product $\iiprod{\sigma,\rho}$.

\textbf{Step~2 ($R[\fa;\rho]$).}
To show a discretization $\fc$ of the reconstruction formula converges to $\fa$ in $C(K)$, it is convenient to regard the integration $\int_{\RR^m\times\RR} [\cdots] \dd\aa\dd b$ in $S$ as the Bochner integral, and the integrand $\gamma(\aa,b)\sigma(\aa\cdot\xx-b)$ as a vector-valued function from $\RR^m\times\RR$ to $C(K)$.

Since $\fa$ is $C^\infty$-smooth, $R[\fa;\rho](\aa,b)$ is bounded and decays rapidly in $\aa$, and thus $R[\fa;\rho]\sigma(\aa\cdot\xx-b)$ is Bochner integrable, that is, 
\begin{align}
    \int_{\RR^m \times \RR} \sup_{\xx \in K} \big| R[\fa;\rho](\aa,b)\sigma(\aa\cdot\xx-b) \big| \dd\aa\dd b < \infty.
\end{align}
To see this, 
the decay property is estimated as follows. For any positive numbers $s,t>1$,
\begin{align}
| R[\fa;\rho](\aa,b) |
&= \frac{1}{2\pi} \Bigg| \int_{\RR} \widehat{\fa}(\omega\aa) \overline{\rho^\sharp(\omega)} e^{i\omega b}\dd\omega \Bigg| \notag\\
&= \frac{1}{2\pi} \Bigg| \int_{\RR} \iprod{\omega\aa}^s \iprod{\omega\aa}^{-s} \iprod{b}^{t} \iprod{b}^{-t} \widehat{\fa}(\omega\aa)  \overline{\rho^\sharp(\omega)}  e^{i\omega b}\dd\omega \Bigg| \notag \\
&\le \frac{1}{2\pi} \Bigg| \int_{\RR} \iprod{\omega\aa}^s    \widehat{\fa}(\omega\aa)  \iprod{\omega}^{-s} \overline{\rho^\sharp(\omega)} \iprod{\triangle_\omega}^{t} e^{i\omega b}\dd\omega \Bigg| \iprod{\aa}^{-s} \iprod{b}^{-t}, %
\end{align}
which asserts the integrability as below
\begin{align}
    \int_{\RR^m\times\RR} \sup_{\xx\in K} | R[\fa;\rho](\aa,b) \sigma(\aa\cdot\xx-b) |\dd\aa\dd b &\lesssim \int_{\RR^m\times\RR} \iprod{\aa}^{-s} \iprod{b}^{-t} \dd \aa \dd b < \infty.
\end{align}

\paragraph{Step~3 ($\fa \sim \fb \sim \fc$).}
Next, take a compact domain ($m+1$-dimensional hypercube) $\cube := \{ (\aa,b) \in \RR^m \times \RR \mid |a_i| \le \delta/2, |b| \le \delta/2 \}$, and put a band-limited function
\begin{align}
    \fb(\xx) := \int_{\cube} R[\fa;\rho](\aa,b) \sigma(\aa\cdot\xx-b) \dd \aa \dd b,
\end{align}
so that $\| \fa - \fb \|_{C(K)} < \eps/3$ (by letting $\delta$ sufficiently large).
Then, let $\cube = \bigsqcup_{i \in I_n} \cube_{ni}$ be a decomposition of the domain $\cube$ into the union of disjoint family of $|I_n| = n^{m+1}$ cubes with volume $\vol(\cube_n) = (\delta/n)^{m+1}$ and the longest diagonal $d_n = \sqrt{m+1}\delta/n$. From each cube, take a point $(\aa_{ni},b_{ni}) \in \cube_{ni}$ as a center of gravity, that is, so that $c_{ni} = \int_{\cube_{ni}} R[\fa;\rho](\aa,b) \dd\aa\dd b = R[\fa;\rho](\aa_{ni},b_{ni}) \vol(\cube_n)$, and put $w_{ni} := R[\fa;\rho](\aa_{ni},b_{ni})$, then put a finite network as
\begin{align}
    \fc(\xx) := \sum_{i \in I_n} c_{ni} \sigma(\aa_{ni}\cdot\xx-b_{ni}).
\end{align}

\paragraph{Step~4 ($\fb \sim \fc$).}
We show $\fc \to \fb$ in $C(K)$.
First, the integrands converge to the limit at almost every $(\aa,b) \in \cube_{ni}$ as
\begin{align}
&\sup_{\xx \in K} \Big| R[\fa;\rho](\aa,b)\sigma(\aa\cdot\xx-b) - w_{ni}\sigma(\aa_{ni}\cdot\xx-b_{ni}) \Big| \\ 
&\le \sup_{\xx \in K} \Big| R[\fa;\rho](\aa,b) \Big| \Big|\sigma(\aa\cdot\xx-b) - \sigma(\aa_{ni}\cdot\xx-b_{ni}) \Big| \notag \\
&\qquad + \Big| R[\fa;\rho](\aa_{ni},b_{ni}) - R[\fa;\rho](\aa,b) \Big| \Big| \sigma(\aa_{ni}\cdot\xx-b_{ni}) \Big| \\
&\le \| R[\fa;\rho] \|_{\infty} \lip(\sigma) \sup_{\xx \in K} \Big| (\aa-\aa_{ni})\cdot\xx - (b-b_{ni})\Big| \notag\\
&\qquad+ \lip(R[\fa;\rho]) \Big|( \aa-\aa_{ni},b-b_{ni} )\Big| \Big| \| \sigma \|_{L^\infty(\RR)} = O(\delta/n) \to 0 \quad n \to \infty.
\end{align}
Besides, the integrands are uniformly bounded as
\begin{align}
    \sup_{\xx \in K} \Big| w_{ni} \sigma(\aa_{ni}\cdot\xx-b_{ni}) \Big| \le \Big| R[\fa;\rho](\aa,b) \Big| \| \sigma \|_{L^\infty(\RR)}, \mbox{ for a.e. } (\aa,b) \in \cube_{ni}.
\end{align}
Therefore, by the dominated convergence theorem for the Bochner integral, we have
\begin{align}
    \| \fb - \fc \|_{C(K)}
    &= \sup_{\xx \in K} \Bigg| \sum_{i \in I_n} \int_{\cube_{ni}} R[\fa;\rho](\aa,b)\sigma(\aa\cdot\xx-b) \dd\aa\dd b - \sum_{i \in I_n} c_{ni} \sigma(\aa_{ni}\cdot\xx-b_{ni})\Bigg| \\
    &\le \sum_{i \in I_n} \int_{\cube_{ni}} \sup_{\xx \in K} \Big| R[\fa;\rho](\aa,b)\sigma(\aa\cdot\xx-b) - w_{ni}\sigma(\aa_{ni}\cdot\xx-b_{ni}) \Big| \dd\aa\dd b\\
    &\to 0, \quad n \to \infty.
\end{align}
Hence by letting $n$ sufficiently large, we have $\| \fc - \fb \|_{C(K)} < \eps/3$. 

To sum up, we have shonw the $cc$-universality:
\begin{align}
    \| f - \fc \|_{C(K)} \le \| f - \fa \|_{C(K)} + \| \fa - \fb \|_{C(K)} + \| \fb - \fc \|_{C(K)} < \eps.
\end{align}
\end{proof}

\paragraph{Notes.}
In the proof, we employed a naive discretization based on the regular grids in $\cube$. However, since we know the closed-form expression of the ridgelet transform, we can discretize it more effectively.
For example, a better discretization scheme is investigated in the so-called \emph{Maurey-Jones-Barron (MJB) theory} and the dimension independent \emph{Barron's bound} (\citep[see, e.g.,][]{kainen.survey}).

\end{document}